\documentclass[10pt,twocolumn,letterpaper]{article}

\usepackage{cvpr}              

\usepackage[dvipsnames]{xcolor}

\definecolor{cvprblue}{rgb}{0.21,0.49,0.74}
\definecolor{greentext}{rgb}{0.0,0.40,0.0}
\usepackage[pagebackref,breaklinks,colorlinks,citecolor=cvprblue]{hyperref}
\usepackage{epsfig}
\usepackage{graphicx}
\usepackage{amsmath}
\usepackage{amssymb}
\usepackage{enumitem}
\usepackage{multicol}
\usepackage{multirow}
\usepackage{booktabs}
\usepackage{pifont}
\usepackage{color, colortbl}
\usepackage{subcaption,amsfonts,dcolumn}

\newcommand\mypara[1]{\vspace{1mm}\noindent\textbf{#1}.}
\newcommand{\cmark}{\ding{51}}%

\definecolor{Gray}{gray}{0.9}

\def\confName{CVPR}
\def\confYear{2024}

\title{Composing Object Relations and Attributes for Image-Text Matching}

\author{
Khoi Pham$^1$ \quad Chuong Huynh$^1$ \quad Ser-Nam Lim$^2$ \quad Abhinav Shrivastava$^1$  \\
\hfill$^1$University of Maryland, College Park\hfill
  $^2$University of Central Florida\hfill\mbox{ }\\
}

\begin{document}
\maketitle
\begin{abstract}
We study the visual semantic embedding problem for image-text matching. Most existing work utilizes a tailored cross-attention mechanism to perform local alignment across the two image and text modalities. This is computationally expensive, even though it is more powerful than the unimodal dual-encoder approach. 
This work introduces a dual-encoder image-text matching model, leveraging a scene graph to represent captions with nodes for objects and attributes interconnected by relational edges. Utilizing a graph attention network, our model efficiently encodes object-attribute and object-object semantic relations, resulting in a robust and fast-performing system.
Representing caption as a scene graph offers the ability to utilize the strong relational inductive bias of graph neural networks to learn object-attribute and object-object relations effectively. To train the model, we propose losses that align the image and caption both at the holistic level (image-caption) and the local level (image-object entity), which we show is key to the success of the model. Our model is termed \textbf{C}omposition model for \textbf{O}bject \textbf{R}elations and \textbf{A}ttributes, \textbf{CORA}. Experimental results on two prominent image-text retrieval benchmarks, Flickr30K and MS-COCO, demonstrate that CORA outperforms existing state-of-the-art computationally expensive cross-attention methods regarding recall score while achieving fast computation speed of the dual encoder. Our code is available at \url{https://github.com/vkhoi/cora_cvpr24}
\end{abstract}    
\section{Introduction}
\label{sec:intro}

\begin{figure}[t]
\centering
\includegraphics[width=1.0\linewidth]{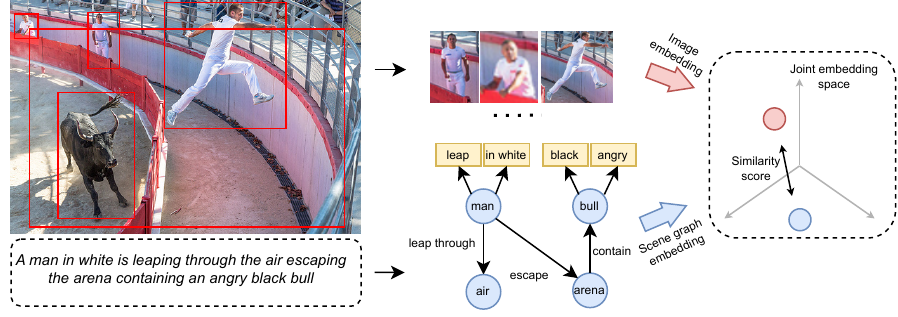}
    \caption{\textbf{Illustration of CORA.} CORA has a dual-encoder architecture, consisting of one encoder that embeds the input image and one encoder that embeds the text caption scene graph into a joint embedding space. (Best viewed in color and zoomed in.)}
\label{fig:teaser}
\vspace{-1em}
\end{figure}

Image-text matching is a fundamental computer vision problem that aims to measure the semantic correspondence between an image and a text. Such correspondence can be used for image retrieval given a text description, or text retrieval provided an image query, both of which are important in various computer vision applications (\eg, weakly supervised problems ~\cite{huynh2023simpson,huynh2024maggie}). The problem is inherently challenging due to the ambiguous nature of the image and text modalities~\cite{song2019polysemous,chun2021probabilistic}. For example, an image can depict a complicated situation that a multitude of different captions can describe, whereas a single caption is too abstract and can semantically apply to multiple images. Various studies have been proposed and can be categorized into two main directions: (1) the unimodal dual encoder and (2) the cross-attention approach.

In the dual-encoder framework, two modality-independent encoders embed the image and text caption separately into a joint embedding space. In this space, a similarity function such as a dot product can measure the image-text similarity. This strategy is also referred to as the global alignment approach, as the goal is to holistically represent an image (or text) as a single embedding. Due to their simplicity and low computational cost (\eg, retrieving an image given a text query can be done via a vector-matrix multiplication with the cached embeddings), such methods are more widely adopted for real-world retrieval databases.

The second approach, cross-attention network, constitutes the majority of recent work. Instead of embedding each modality separately, cross-modality attention is adopted to locally align fine-grained visual cues of an image (image regions) with textual cues of a caption (word tokens), from which the overall correspondence score is aggregated. 
While this approach outperforms dual encoder in terms of power, it presents a substantial computational challenge. Upon receiving a text (or image) query, every image vs. text query pair must be processed through the cross-attention model to determine their similarity scores. This requirement renders the method impractical for retrieval systems managing large databases due to its extensive computational demands.
This work focuses on the dual-encoder approach and shows that our dual-encoder proposal even outperforms the SOTA cross-attention networks.

Existing approaches use a text sequence model (\eg, GRU~\cite{chung2014empirical}, LSTM~\cite{hochreiter1997long}) to encode the text caption. A text usually contains an extensive range of semantic information, such as object categories, attributes of objects, and relations between objects. Attributes describe appearance of objects~\cite{patterson2016coco,krishna2017visual,pham2021learning,saini2022disentangling,pham2022improving}, while relations describe how objects interact with one another~\cite{zellers2018neural}. Forcing a text sequence model to learn to parse a caption into different levels of semantics is challenging, especially in the low data regime. For example, by design, a sequence model that simply processes a caption from left to right (GRU, LSTM) may find it challenging to determine which attributes belong to an object and which objects participate in a relation. Numerous works have shown that Transformer-based text sequence models (BERT~\cite{devlin2018bert}) can produce good structural parsing of a sentence \cite{hewitt2019structural}, however, these models must be trained on large amounts of data. Nevertheless, it has been shown in~\cite{chefer2023attend} that even the CLIP text encoder~\cite{radford2021learning} in Stable Diffusion~\cite{rombach2022high,phung2024attenref} still exhibits incorrect object-attribute binding (\ie, pair an attribute with the wrong object in the sentence) despite having been trained on large datasets. Therefore, it becomes desirable to have a text embedding model that can capture the semantic relations between concepts accurately.

In this work, instead of a sequence model, we propose representing a caption as a scene graph of object and attribute nodes connected by relation edges. An example of a scene graph is illustrated in Fig.~\ref{fig:teaser}, where we show that semantic structures such as object-attribute and object-object pairings are already organized. To this end, we propose our \textbf{C}omposition model for \textbf{O}bject \textbf{R}elations and \textbf{A}ttributes, \textbf{CORA}, a dual-encoder model for image-text matching. On the image side, we re-use GPO~\cite{chen2021learning} which is a SOTA pooling operator for image-text matching to embed the image as a vector. On the text side, we propose to use a graph attention network~\cite{velivckovic2017graph, brody2021attentive} with strong relational inductive bias to produce a holistic scene graph embedding for the caption. Scene graph-based approaches have been previously explored in~\cite{wang2020cross, li2020visual, liu2020graph, long2022gradual} for image-text matching, but they all employ expensive cross-attention. In addition to the margin-based triplet ranking loss~\cite{faghri2017vse++} adopted by prior work, we propose a contrastive loss to guide CORA in making alignment at both the holistic image-caption level and the local image-object entity level. The proposed loss helps make training more stable and result in better downstream retrieval accuracy, as well as additionally acquires CORA with the image-object entity retrieval capability.

Our model is evaluated on two image-text retrieval benchmarks, Flickr30K and MS-COCO, where it outperforms SOTA dual-encoder and expensive cross-attention methods. Our paper makes the following contributions:
\begin{itemize}[noitemsep,left=0pt]
    \item We propose CORA, a dual encoder for image-text matching that uses a graph attention network instead of a sequence model to produce scene graph embedding for a caption.
    \item We propose using contrastive loss that trains the model to make global alignment (image-caption) and local alignment (image-object entity), resulting in more stable training, better retrieval accuracy, and image-object retrieval capability.
    \item Our model CORA achieves SOTA retrieval performance on Flickr30K and MS-COCO, two prominent benchmarks for image-text retrieval.
\end{itemize}
\section{Related Work}
\label{sec:related_work}

\mypara{Dual-encoder} This approach is dominant in earlier works~\cite{frome2013devise, kiros2014unifying, wang2016learning, faghri2017vse++, li2019visual} in image-text matching. The image and text captions are independently embedded in a joint metric space where matching image-caption pairs are located close to each other.
Existing work in this paradigm often improves the joint embedding space by introducing new losses~\cite{faghri2017vse++,chun2021probabilistic}, proposing new architecture for each modality encoder~\cite{li2019visual,wen2020learning,wu2019learning}, or learning better pooling methods~\cite{chen2021learning,li2022multi}.
For example, VSE++~\cite{faghri2017vse++} proposes a triplet loss with hard negative mining which has been adopted by all following image-text matching work. VSRN~\cite{li2019visual}, DSRAN~\cite{wen2020learning}, SAEM~\cite{wu2019learning} implement graph convolution and self-attention to improve the encoder architecture.
GPO~\cite{chen2021learning} achieves competitive results by designing a new pooling operator that can learn from data. Recently, MV-VSE~\cite{li2022multi} and SDE~\cite{kim2023improving} propose using multiple embeddings per sample data, and HREM~\cite{fu2023learning} presents a dual-encoder model that can be trained with a cross-modality matching loss for enhancing the embedding quality.

\mypara{Cross-attention} In contrast to embedding the image and text independently, this approach considers the fine-grained local correspondence between image features and text tokens before computing the similarity. SCAN~\cite{lee2018stacked} is the first representative work that introduces this idea of using cross-attention between the two modalities to find their alignments. CAAN~\cite{zhang2020context} later improves the idea by employing an additional intra-modal interaction step after the cross-modal interaction. SGARF~\cite{diao2021similarity} proposes to learn jointly from both the global and local alignment to highlight important image regions.
Recently, NAAF~\cite{zhang2022negative} encourages the dissimilarity degrees between mismatched pairs of image region and word to boost the similarity matching, and CHAN~\cite{pan2023fine} proposes a new cross-modal alignment method that can neglect the redundant misalignments. 

\mypara{Graph-based image-text matching} Among both dual-encoder and cross-attention methods, some have utilized scene graphs as part of their pipeline for more accurate image-text alignment~\cite{wang2020cross,li2020visual,liu2020graph,long2022gradual}. Frameworks based on this approach leverage the capacity of Graph Convolutional Networks (GCN) to capture the spatial and semantic relationships between visual regions and textual tokens.
For example, SGM~\cite{wang2020cross}, GCN+DIST~\cite{li2020visual}, GraDual~\cite{long2022gradual} utilize off-the-shelf visual scene graph generator~\cite{zellers2018neural} to extract scene graph from images, then perform cross-modal alignment between the visual and textual graph. GSMN~\cite{liu2020graph}, on the other hand, uses a fully connected graph for the visual regions but additionally uses the regions' polar coordinates to encode their spatial relationships.

In our work, we build upon the scene graph representation of the caption to develop the text encoder for our dual-encoder model. Our model focuses on explicitly learning to compose objects with their attributes and all objects in the scene through their relationships to produce a single embedding vector for the text rich in semantic information. To the best of our knowledge, there has yet to be any previous \underline{dual-encoder} work on explicitly capturing the object, attribute, and relation semantics through scene graphs for image-text matching. Our method is different from previous graph-based approaches in that we do not use external visual scene graph generator, which is prone to wrong prediction, and we carefully design a 2-step graph encoding approach trained with a contrastive loss to align at both the global image-text and local image-object level. Our network outperforms SOTA methods without the heavy cross-attention module.
\section{Method}
\label{sec:method}

This section describes our Composition model for Object Relations and Attributes. We first describe the overall framework in~\cref{sec:overall}, then present in \cref{sec:feature_extraction} how we perform visual embedding on the input image, how we parse the text caption into a scene graph and extract text features for each node in the graph. In~\cref{sec:scene_graph_embed}, we describe how we can embed this scene graph into the joint embedding space with the image using the graph attention network. Finally, training objectives are detailed in~\cref{sec:train_objective}.

\subsection{Overall Framework}
\label{sec:overall}
\begin{figure*}
\centering
\includegraphics[width=0.95\linewidth]{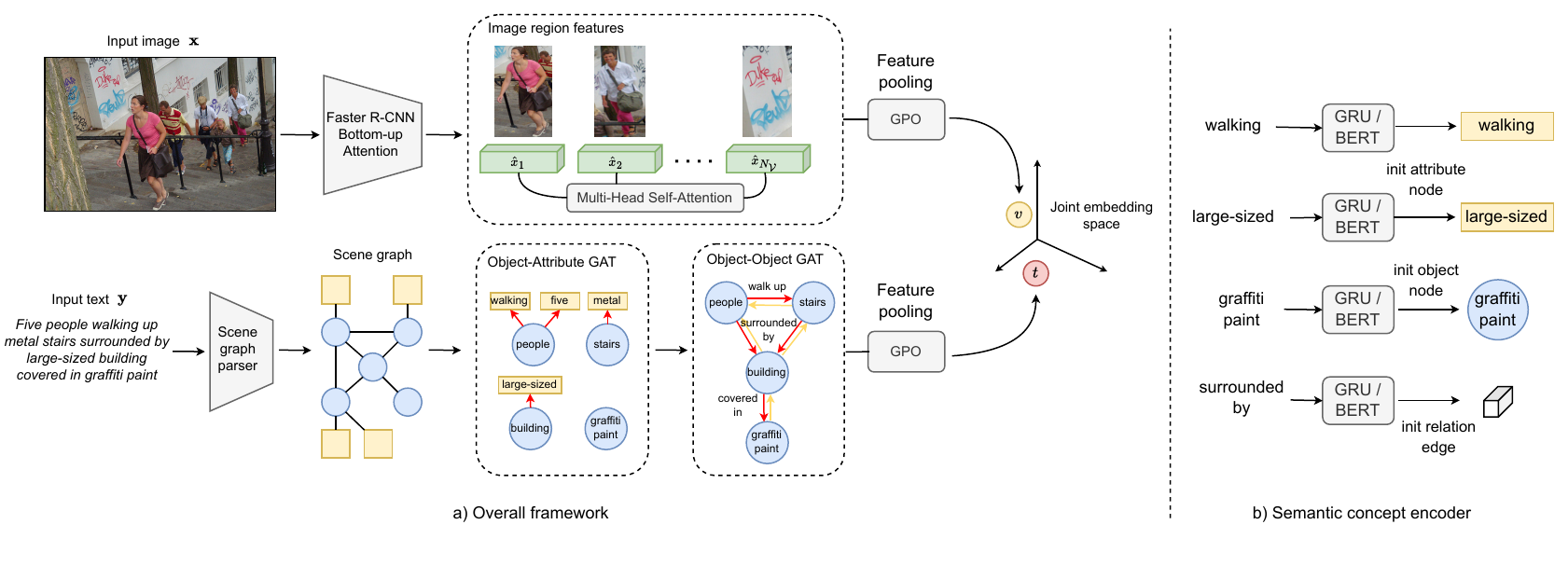}
    \caption{\textbf{Overview of CORA.} a) CORA consists of (1) an image encoder that detects and extracts the salient regions' features from the input image, contextualizes them through a multi-head self-attention, then aggregates them into a single image embedding through the GPO~\cite{chen2021learning} pooling operator, (2) a text encoder that first parses the input text into a scene graph where all semantic information is readily organized, then two graph attention networks Object-Attribute GAT and Object-Object GAT are used to encode this graph into the same joint space with the image. The red arrow denotes the edge of the active role, while the yellow arrow is for the passive role in the relation (refer to~\cref{sec:scene_gat}). b) The semantic concept encoder that uses GRU or BERT to encode each semantic concept in the graph corresponding to the object, attribute nodes and relation edges. }
\label{fig:model}
\end{figure*}

We begin by describing the overall framework of CORA, which is illustrated in~\cref{fig:model}. The model consists of two encoders: a visual encoder $f^\mathcal{V}$ that takes in an input image $\mathbf{x}$ and produces the image embedding vector $v = f^\mathcal{V}(\mathbf{x}) \in \mathbb{R}^D$; and a text encoder $f^\mathcal{T}$ that takes in the text caption $\mathbf{y}$ and produces its embedding $t = f^\mathcal{T}(\mathbf{y}) \in \mathbb{R}^D$ in the joint $D$-dimensional embedding space. Instead of embedding the text caption directly, we first parse it into a scene graph using a parser $\phi^\text{SG}$, then apply a graph attention network $f^\mathcal{G}$ to embed this scene graph. Our text embedding formulation therefore can be rewritten as $t = f^\mathcal{G}(\phi^\text{SG}(\mathbf{y}))$.

The similarity score between the image and the text caption is defined as the cosine similarity between their embeddings $v$ and $t$:
\begin{equation}
    \text{sim}\mathbf{(x, y)} = \frac{v^\text{T}t}{\|v\|\|t\|}.
\end{equation}

The dual-encoder is efficient for image-text retrieval. In the context of image retrieval, all image embeddings can be computed and cached in advance. When a text query arrives, it only needs to be embedded with $f^\mathcal{G}(\phi^\text{SG}(.))$, then a simple vector-matrix multiplication is sufficient to retrieve all nearest neighbor images of the query.

\subsection{Feature Extraction}
\label{sec:feature_extraction}

\mypara{Visual feature extractor} Given an input image $\mathbf{x}$, we follow convention from prior work to use the pre-trained bottom-up detection model BUTD~\cite{anderson2018bottom}. With this model, the top-36 most confident salient regions in $\mathbf{x}$ are detected, along with their visual features $\{x_k \in \mathbb{R}^{2048}\}_{k=1}^{N_\mathcal{V}}, N_\mathcal{V}=36$. The detection model used here is a Faster R-CNN with ResNet-101 backbone \cite{he2016deep}, pre-trained on Visual Genome \cite{krishna2017visual}. We also transform the region features with an FC layer so that they have the same dimensions as the joint embedding space: $x_k \in \mathbb{R}^D$. Furthermore, we also apply multi-head self-attention to contextualize the region features against one another. Then, in order to perform feature aggregation on this set to obtain a holistic representation for the input image $v = f^\mathcal{V}(\mathbf{x}) \in \mathbb{R}^D$, we implement $f^\mathcal{V}$ using GPO~\cite{chen2021learning} which is a SOTA pooling operator for image-text matching. Essentially, GPO learns to generate the best pooling coefficient for every visual region, which is better than naively applying mean pooling over the visual feature set.

\mypara{Scene graph parser} Formally, we implement a textual scene graph parser that can construct a graph $G = (V, E)$ given a text caption $\mathbf{y}$, where $V = O \cup A$ denotes the set of object nodes $O$ and attribute nodes $A$, and $E = E_\text{OA} \cup E_\text{OO}$ represents the set of object-attribute edges $E_\text{OA}$ and object-object relation edges $E_\text{OO}$. Example of a scene graph is illustrated in~\cref{fig:teaser}. We implement a scene graph parser based on \cite{schuster2015generating,wu2019unified}, using the syntactical dependency parser from the spaCy library \cite{spacy}. We develop rules to extract object nouns (\eg, \textit{construction worker}), adjective and verb attributes (\eg, \textit{salmon-colored, sitting}), verb relations (\eg, \textit{person-\underline{jump over}-fence, dog-\underline{wear}-costume}), and preposition relations (\eg, \textit{flag-\underline{above}-building}). Existing scene graph parsers \cite{schuster2015generating,wu2019unified} are developed upon inferior language toolkits, thus often misdetect concepts (\eg, those consisting of multiple word tokens are not detected). The implementation of our parser is made publicly available.

\mypara{Semantic concept encoder} We denote the set of object nodes $O = \{o_i\}$, attribute nodes $A = \{a_i\}$, and object-object relation edges $E_\text{OO} = \{r_{ij}\}$. These concepts are still in text format that need to be encoded into vector representation. As these concepts often consist of multiple word tokens (\eg, \textit{pair of shoes, jump over}), we use a text sequence model as a phrase encoder to encode all semantic concepts. To demonstrate the generalizability of our method across different language features, we implement this semantic concept encoder using Bi-GRU~\cite{chung2014empirical} and BERT~\cite{devlin2018bert}. For Bi-GRU, given an $L$-word semantic concept, we use the GloVe~\cite{pennington2014glove} word embedding of each word to obtain a sequence of $L$ $300$-dimensional vectors. Next, we employ a Bi-GRU and take the final hidden states as the representation for the concept $c \in \mathbb{R}^{300}$. For BERT, we use the average of the output hidden states of all tokens at the last layer to represent the concept $c \in \mathbb{R}^{768}$. For both types of features, we then use an FC layer to transform the concept embedding to have the same dimension $D$ as the joint embedding space. These concept embeddings are used to initialize the node features for $\{o_i\}$ and $\{a_i\}$ and the edge features for $\{r_{ij}\}$ in the scene graph.

\subsection{Scene Graph Embedding}
\label{sec:scene_graph_embed}

After obtaining the graph structure from the parser and the initialized features for all nodes and edges in the graph, we continue to elaborate on our scene graph embedding method as follows. The core idea of our method is that the scene semantics should be composed at two levels in a bottom-up manner, where we use a separate graph attention network (GAT) \cite{velivckovic2017graph, brody2021attentive} for each level. At the bottom level, a GAT models the relations between an object and its associated attributes. At the top level, another GAT is used to model the relations between solely the objects, compose them together and produce the final scene embedding.

\mypara{GAT Preliminaries} GAT is among the most popular graph neural network methods, with SOTA results in graph representation learning. We follow the implementation of GATv2 \cite{brody2021attentive}, which is an improved version of the original GAT \cite{velivckovic2017graph}. We provide a brief description of GATv2 here. Given a directed graph $G = (V, E)$, containing nodes $V = \{1, ..., N\}$ and $E \subseteq V \times V$ where $(j, i) \in E$ denotes an edge from node $j$ to $i$. For each node $i$, we also have its initial representation denoted as $h_i \in \mathbb{R}^d$. In a message passing step, to update features for node $i$, we first compute the importance value of neighbor node $j$ w.r.t. $i$ as following
\begin{equation}
    e(h_i, h_j) = a^\text{T}\text{LeakyReLU}(W \cdot [h_i \| h_j]),
\end{equation}
where $\|$ denotes vector concatenation, $W \in \mathbb{R}^{d \times 2d}, a \in \mathbb{R}^{d \times 1}$. Followed by softmax, normalized attention coefficients of all neighbors $j \in \mathcal{N}_i$ can be obtained: $\alpha_{i,j} = \text{softmax}(e(h_i, h_j))$. Then, new representation $h_i$ for node $i$ is aggregated by
\begin{equation}
    h'_i = \text{ReLU}(\sum_{j\in \mathcal{N}_i} \alpha_{i,j} Wh_j).
\end{equation}
Formally, the output of one GAT layer on a graph $G$ is
\begin{equation}
    \{h'_i\} = \text{GAT}(\{h_i\}, G).
\end{equation}

\subsubsection{Object-Attribute GAT}
\label{subsec:objatt_gat}
At the bottom level, we care about how the semantic representation of an object is modified by its connected attributes in the graph. These attributes are modifiers that alter the visual appearance of the object. Because an attribute of one object should in no way alter the appearance of another object, in this step, we apply GAT only on the subgraph $G_\text{OA} = (V, E_\text{OA})$ consists of only edges between the object and attribute nodes.

We denote $\{h_i\}_{i=1}^{|V|}, h_i \in \mathbb{R}^D$ as the initial representations for all nodes in the graph. These representations are initialized from the aforementioned semantic concept embedding step. We train a graph attention network, which we name $\text{GAT}_\text{Obj-Att}$ to perform message passing in graph $G_\text{OA}$. The updated representation of all nodes is therefore
\begin{equation}
    \{h'_i\} = \text{GAT}_\text{Obj-Att}(\{h_i\}, G_\text{OA}).
\end{equation}

At the output, we are only interested in the updated representation of the set of object nodes. Since these objects have been composed with their corresponding attributes, we name them as \textbf{entities} and denote them as $\{e_i\}_{i=1}^{|O|}$, which will be used in one of our proposed losses.

\subsubsection{Object-Object Relation GAT}
\label{sec:scene_gat}
At the top level, after acquiring the entity embeddings $\{e_i\}_{i=1}^{|O|}$ for all object nodes, we continue to apply another GAT, which we name $\text{GAT}_\text{Obj-Obj}$ on the subgraph $G_\text{OO} = (O, E_\text{OO})$ consisting of only object nodes and edges between them. Because these object nodes are connected with object-object relation edges $\{r_{ij}\}$, our first step before applying GAT is to contextualize the entity embeddings with their corresponding edges.

\mypara{Edge features} Consider a directed relation edge $r_{ij}$. In this relation, node $i$ plays the subject (active) role while node $j$ plays the object (passive) role. For example, in the relation \textit{man-hold-cup}, \textit{man} is the subject while \textit{cup} is the object. To obtain the edge features for this relation, we concatenate its semantic encoding $r_{ij}$ with the embedding of the entity that plays the passive role $e_j$ as follows: $r'_{ij} = [r_{ij} \| e_j]$. While existing work \cite{nguyen2021defense} often concatenates $r_{ij}$ with both the subject and object entity, in our work, we find that it is empirically better to characterize a relation with only the passive object entity. This is intuitively reasonable since the meaning of a relation such as \textit{hold-cup, use-computer} does not depend on what kind of subject is involved.

\mypara{Edge-contextualized entity features} Consider object node $i$, we define $\text{Active}(i) = \{j | r_{ij}\in E_\text{OO}\}$ consisting all nodes that node $i$ has a subject (active) relation with. Vice-versa, we define $\text{Passive}(i) = \{j | r_{ji}\in E_\text{OO}\}$ which is all nodes that node $i$ has an object (passive) relation. We contextualize the embedding of entity $i$ with its edges as
\begin{equation}
    e'_i = e_i + \frac{\sum_{j\in \text{Active}(i)}W_Ar'_{ij}}{|\text{Active}(i)|} + \frac{\sum_{j\in \text{Passive}(i)}W_Pr'_{ji}}{|\text{Passive}(i)|},
\end{equation}
where $W_A$ and $W_P$ are two learnable matrices mapping edge features to have the same dimension with entity embeddings.

\mypara{Scene graph embedding} With $\{e'_i\}_{i=1}^{|O|}$ as the initial representation for all object nodes. We train a $\text{GAT}_\text{Obj-Obj}$ on graph $G_\text{OO}$. The updated representation for all nodes is
\begin{equation}
    \{\hat{e}_i\} = \text{GAT}_\text{Obj-Obj}(\{e'_i\}, G_\text{OO}).
\end{equation}
In order to pool the whole graph into one single embedding vector, we also use GPO~\cite{chen2021learning} similar to our visual feature extraction step. We take the output representation that is pooled from GPO as the scene embedding $t$ to represent the original input text caption in the joint embedding space.

\subsection{Training Objectives}
\label{sec:train_objective}
Let $B = \{(v_i, t_i, \{e_{ik}\}_{k=1}^{|O_i|})\}_{i=1}^N$ be the training batch of output image embedding $v_i$ of the $i$-th image, output text embedding $t_i$ of the $i$-th text caption from $\text{GAT}_\text{Obj-Obj}$, and set of output entity embeddings $\{e_{ik}\}_{k=1}^{|O_i|}$ of the $i$-th text caption from $\text{GAT}_\text{Obj-Att}$. It is reminded that these entities $\{e_{ik}\}$ are embeddings of the object nodes in the scene graph of $t_i$. We train our model CORA with the following losses. For brevity, we denote $s(v, t) = v^\text{T}t / (\|v\| \|t\|)$ to be the cosine similarity between $v$ and $t$.

\mypara{Triplet loss with hardest negatives} Following prior work in image-text retrieval \cite{faghri2017vse++, chen2021learning}, we also adopt the hinge-based triplet loss with hardest negative mining,
\begin{align}
    \mathcal{L}_\text{HARD} = \sum_{i} \max_j [\alpha + s(v_i, t_j) - s(v_i, t_i)]_+ \\
    + \max_j [\alpha + s(v_j, t_i) - s(v_i, t_i)]_+.
\end{align}

Essentially, for every matching image-caption $v_i$ and $t_i$ in the training batch, this loss looks for the negative caption $t_j$ that is closest to $v_i$, and the negative image $v_j$ that is closest to $t_i$ in the embedding space. $t_j$ and $v_j$ are the hardest negatives in the training batch and help provide a strong discriminative learning signal to the model.

\mypara{Contrastive loss} As observed by previous work \cite{chen2021learning}, the hardest triplet loss above results in unstable learning during early training epochs. We find that applying a contrastive loss that encourages the model to align the output representations of all matching image, text, and object entity together results in more stable training and better final results. Because the entity embeddings $\{e_{ik}\}_{k=1}^{|O_i|}$ are also involved in the equation here, our model CORA is also trained to perform image retrieval given an object entity (\eg, image searching for \textit{straw hat}). The loss is formulated as follows
\begin{align}
    \mathcal{L}_\text{CON}  = -\sum_{i} \sum_{u} \log \frac{\exp{(s(v_i, u)})}{\sum_{u' \in \mathcal{N}_i} \exp{(s(v_i, u'))}} \\
    - \sum_{i} \sum_{u} \log \frac{\exp{(s(v_i, u)})}{\sum_{v' \in \mathcal{N}_u} \exp{(s(v', u))}},
\end{align}
where $u \in \{ t_i \} \cup \{e_{ik}\}_{k=1}^{|O_i|}$ is the semantic embedding of either the text or an object entity corresponding to image $i$, $\mathcal{N}_i$ is the negative set of semantic concepts that do not correspond to image $i$, and similarly $\mathcal{N}_u$ is the negative set of images that do not contain semantic concept $u$.

\mypara{Specificity loss} The contrastive loss above aligns the embeddings of image, text, and entity together in the joint space. In addition, we would like to impose some structure in this space such that the similarity between an image $v_i$ and text $t_i$ should be larger than between $v_i$ and all entities $\{e_{ik}\}$. The reason is that a caption always depicts more semantic information than an entity alone, hence $t_i$ should be more specific w.r.t. $v_i$ and exhibits a larger similarity score. The loss takes the form of a hinge-based triplet loss
\begin{equation}
    \mathcal{L}_\text{SPEC} = \sum_i \sum_k [\alpha + s(v_i, e_{ik}) - s(v_i, t_i)]_+.
\end{equation}

Our overall loss is therefore a weighted sum of all losses:
\begin{equation}
    \mathcal{L} = \mathcal{L}_\text{HARD} + \lambda_\text{CON}\mathcal{L}_\text{CON} + \lambda_\text{SPEC}\mathcal{L}_\text{SPEC}.
\end{equation}
\section{Experiments}
\label{sec:experiments}

We describe our experiments to validate the effectiveness of CORA. We describe the datasets in Sec~\ref{sec:dataset} and analyze the results in Sec~\ref{sec:results}. To validate design choices, we present ablations in Sec~\ref{sec:ablation}. We refer to supplementary for implementation, qualitative results, and inference time analysis.

\subsection{Dataset and Evaluation Metrics}
\label{sec:dataset}
\mypara{Datasets} We perform experiments on two standard benchmarks, Flickr30K \cite{plummer2015flickr30k} and MS-COCO \cite{lin2014microsoft}, on the image-to-text retrieval (I2T) and text-to-image retrieval (T2I) tasks. In both datasets, every image is annotated with five text descriptions. As in prior work \cite{chen2021learning}, we follow the splits convention on both datasets. Flickr30K contains 31K images, of which 29K images are for training, 1K for validation, and 1K for testing. MS-COCO provides 123,287 images and is split into 113,287 images for training, 5000 images for validation, and 5000 images for testing.

\mypara{Metrics} We report the commonly used Recall@K (R@K), where K $\in \{1, 5, 10\}$. This metric computes the percentage of queries where the correct match appears in the top-K retrievals. To summarize performance, we report RSUM which is the sum of R@K at all values of K $\in \{1, 5, 10\}$ on I2T and T2I tasks. For MS-COCO, by convention, the results are reported in two settings: 5K setting, and 1K setting where the results are averaged over five 1K data folds.

\subsection{Quantitative Results}
\label{sec:results}

\begin{table}[t]
\centering
\caption{\textbf{Our framework achieves the best and second-best on the Flickr30K dataset with two different encoders.} Without the CA - ``cross-attention'', our method still has competitive results to other baselines. $\dagger$ denotes methods that use ensembling of multiple models, and we highlight \textbf{the highest} and \underline{second-highest} RSUM for each section.}
\scalebox{0.6}{
    \begin{tabular}{@{}l|c|c|ccc|ccc|c@{}}
    \toprule
    
    \multirow{2}{*}{Method}
    & \multirow{2}{*}{Venue}
    & \multirow{2}{*}{CA}
    & \multicolumn{3}{c|}{Image $\rightarrow$ Text} 
    & \multicolumn{3}{c|}{Text $\rightarrow$ Image} 
    & \multirow{2}{*}{RSUM} \\
    \cmidrule{4-9}
    
    & & & R@1 & R@5 & R@10 & R@1 & R@5 & R@10 & \\ \midrule
    
    \multicolumn{10}{l}{\textit{\textbf{Faster R-CNN + Bi-GRU}}} \\ \midrule

    SCAN$^\dagger$~\cite{lee2018stacked} & ECCV'18 & \cmark & 67.4& 90.3& 95.8& 48.6& 77.7& 85.2& 465.0 \\

    VSRN~\cite{li2019visual} & ICCV'19 & & 71.3& 90.6& 96.0& 54.7& 81.8& 88.2& 482.6 \\

    SGM~\cite{wang2020cross} & WACV'20 & \cmark & 71.8 & 91.7 & 95.5 & 53.5 & 79.6 & 86.5 & 478.6 \\
    
    GCN+DIST~\cite{li2020visual} & CVPR'20 & \cmark & 70.8 & 92.7 & 96.0 & 60.9 & 86.1 & 91.0 & 497.5 \\

    GSMN$^\dagger$~\cite{liu2020graph} & CVPR'20 & \cmark & 76.4 & 94.3 & 97.3 & 57.4 & 82.3 & 89.0 & 496.8 \\

    CAAN~\cite{zhang2020context} & CVPR'20 & \cmark & 70.1 & 91.6 & 97.2 & 52.8 & 79.0 & 87.9 &  478.6 \\

    VSE${_\infty}$~\cite{chen2021learning} & CVPR'21 & & 76.5& 94.2& 97.7& 56.4& 83.4& 89.9& 498.1 \\

    VSRN$_{++}$~\cite{li2019visual} & PAMI'22 & & 79.2 & 94.6 & 97.5 & 60.6 & 85.6 & 91.4 & 508.9 \\

    SGARF$^\dagger$~\cite{diao2021similarity} & AAAI'23 & \cmark & 77.8 & 94.1 & 97.4 & 58.5 & 83.0 & 88.8 & 499.6 \\

    CODER~\cite{wang2022coder} & ECCV'22 & \cmark & 79.4 & 94.9 & 97.7 & 59.0 & 85.2 & 91.0 & 507.2 \\

    MV-VSE$^\dagger$~\cite{li2022multi} & IJCAI'22 & & 79.0 & 94.9 & 97.7 & 59.1 & 84.6 & 90.6 & 505.8 \\

    GraDual$^\dagger$~\cite{long2022gradual} & WACV'23 & \cmark & 78.3 & 96.0 & 98.0 & 60.4 & 86.7 & 92.0 & 511.4 \\

    CHAN~\cite{pan2023fine} & CVPR'23 & \cmark & 79.7 & 94.5 & 97.3 & 60.2 & 85.3 & 90.7 & 507.7 \\

    NAAF$^\dagger$~\cite{zhang2022negative} & CVPR'23 & \cmark & 81.9& 96.1& 98.3& 61.0& 85.3& 90.6& 513.2 \\

    SDE$^\dagger$~\cite{kim2023improving} & CVPR'23 & & 80.9 & 94.7 & 97.6 & 59.4 & 85.6 & 91.1 & 509.3 \\

    HREM$^\dagger$~\cite{fu2023learning} & CVPR'23 & & 81.4 & 96.5 & 98.5 & 60.9 & 85.6 & 91.3 & 514.2 \\

    \rowcolor{Gray}
    \textbf{Ours} & & & 82.2 & 95.6 & 97.7 & 61.8 & 86.5 & 92.0 & \underline{515.8} \\
    \rowcolor{Gray}
    \textbf{Ours}$^\dagger$ & & & 82.3 & 96.1 & 98.0 & 63.0 & 87.4 & 92.8 & \textbf{519.6} \\

    \midrule
    
    \multicolumn{9}{l}{\textit{\textbf{Faster R-CNN + BERT}}} \\ \midrule

    VSE${_\infty}$~\cite{chen2021learning} & CVPR'21 & & 81.7 & 95.4 & 97.6 & 61.4 & 85.9 & 91.5 & 513.5 \\

    CODER~\cite{wang2022coder} & ECCV'22 & \cmark & 83.2 & 96.5 & 98.0 & 63.1 & 87.1 & 93.0 & 520.9 \\

    MV-VSE$^\dagger$~\cite{li2022multi} & IJCAI'22 & & 82.1 & 95.8 & 97.9 & 63.1 & 86.7 & 92.3 & 517.5 \\

    CHAN~\cite{pan2023fine} & CVPR'23 & \cmark & 80.6 & 96.1 & 97.8 & 63.9 & 87.5 & 92.6 & 518.5 \\

    HREM$^\dagger$~\cite{fu2023learning} & CVPR'23 & & 84.0 & 96.1 & 98.6 & 64.4 & 88.0 & 93.1 & \textbf{524.2} \\

    \rowcolor{Gray}
    \textbf{Ours} & & & 83.7 & 96.6 & 98.3 & 62.3 & 87.1 & 92.6 & 520.1 \\
    \rowcolor{Gray}
    \textbf{Ours}$^\dagger$ & & & 83.4 & 95.9 & 98.6 & 64.1 & 88.1 & 93.1 & \underline{523.3} \\
                            
    \bottomrule
    \end{tabular}
}

\vspace{-3mm}
\label{tab:f30k_comparison}
\end{table}

\begin{table*}[t]
\centering
\caption{
\textbf{Our method yields competitive results on the MS-COCO dataset.} Our performance is competitive in all test schema with previous works, especially on the simple Bi-GRU architecture. $\dagger$ denotes methods that use ensembling of multiple models. \textbf{Bold} and \underline{underline} highlight the best and second-best performance.
}
\scalebox{0.88}{
\footnotesize
\setlength{\tabcolsep}{5pt}
\begin{tabular}{@{}l|c|c|ccc|ccc|c|ccc|ccc|c@{}}
\toprule
    \multicolumn{1}{l}{\multirow{3}{*}[-1.3em]{Method}}
    & \multicolumn{1}{l}{\multirow{3}{*}[-1.3em]{\ \ \ Venue}}
    & \multicolumn{1}{c|}{\multirow{3}{*}[-0.65em]{Cross-}} 
    & \multicolumn{7}{c|}{MS-COCO 5-fold 1K Test} 
    & \multicolumn{7}{c}{MS-COCO 5K Test} \\ \midrule
    & & 
    & \multicolumn{3}{c|}{Image $\rightarrow$ Text} 
    & \multicolumn{3}{c|}{Text $\rightarrow$ Image}
    & \multicolumn{1}{c|}{\multirow{2}{*}{\textsc{RSUM}}}
    & \multicolumn{3}{c|}{Image $\rightarrow$ Text} 
    & \multicolumn{3}{c|}{Text $\rightarrow$ Image}
    & \multicolumn{1}{c}{\multirow{2}{*}{RSUM}} \\
    \cmidrule{4-9} \cmidrule{11-16}
    \multicolumn{1}{c|}{}
    & & Attention
    & R@1 & R@5 & R@10 & R@1 & R@5 & R@10 &
    & R@1 & R@5 & R@10 & R@1 & R@5 & R@10 & \\
\midrule %
\multicolumn{16}{l}{\textit{\textbf{Faster R-CNN + Bi-GRU}}} \\ \midrule

SCAN$^\dagger$~\cite{lee2018stacked} & ECCV'18 & \cmark
    & 72.7& 94.8& 98.4& 58.8& 88.4& 94.8& 507.9
    & 50.4& 82.2& 90.0& 38.6& 69.3& 80.4& 410.9 \\

VSRN~\cite{li2019visual} & ICCV'19 &
    & 76.2 & 94.8 & 98.2 & 62.8 & 89.7 & 95.1 & 516.8
    & 53.0 & 81.1 & 89.4 & 40.5 & 70.6 & 81.1 & 415.7 \\

SGM~\cite{wang2020cross} & WACV'20 & \cmark
    & 73.4 & 93.8 & 97.8 & 57.5 & 87.3 & 94.3 & 504.1 
    & 50.0 & 79.3 & 87.9 & 35.3 & 64.9 & 76.5 & 393.9 \\

CAAN~\cite{zhang2020context} & CVPR'20 & \cmark
    & 75.5 & 95.4 & 98.5 & 61.3 & 89.7 & 95.2 & 515.6
    & 52.5 & 83.3 & 90.9 & 41.2 & 70.3 & 82.9 & 421.1 \\

VSE$_\infty$~\cite{chen2021learning} & CVPR'21 &
    & 78.5 & 96.0 & 98.7 & 61.7 & 90.3 & 95.6 & 520.8 
    & 56.6 & 83.6 & 91.4 & 39.3 & 69.9 & 81.1 & 421.9 \\

SGARF$^\dagger$~\cite{diao2021similarity} & AAAI'23 & \cmark
    & 79.6 & 96.2 & 98.5 & 63.2 & 90.7 & 96.1 & 524.3 
    & 57.8 & - & 91.6 & 41.9 & - & 81.3 & - \\

CODER~\cite{wang2022coder} & ECCV'22 & \cmark
    & 78.9 & 95.6 & 98.6 & 62.5 & 90.3 & 95.7 & 521.6
    & 58.5 & 84.3 & 91.5 & 40.9 & 70.8 & 81.4 & 427.4 \\

MV-VSE$^\dagger$~\cite{li2022multi} & IJCAI'22 & 
    & 78.7 & 95.7 & 98.7 & 62.7 & 90.4 & 95.7 & 521.9
    & 56.7 & 84.1 & 91.4 & 40.3 & 70.6 & 81.6 & 424.6 \\

GraDual$^\dagger$~\cite{long2022gradual} &  WACV'23 & \cmark
    & 77.0 & 96.4 & 98.6 & 65.3 & 91.9 & 96.4 & 525.6
    & - & - & - & - & - & - & - \\

CHAN~\cite{pan2023fine} & CVPR'23 & \cmark 
    & 79.7 & 96.7 & 98.7 & 63.8 & 90.4 & 95.8 & 525.1
    & 60.2 & 85.9 & 92.4 & 41.7 & 71.5 & 81.7 & 433.4 \\

NAAF$^\dagger$~\cite{zhang2022negative} & CVPR'23 & \cmark
    & 80.5 & 96.5 & 98.8 & 64.1 & 90.7 & 96.5 & 527.2 
    & 58.9 & 85.2 & 92.0 & 42.5 & 70.9 & 81.4 & 430.9 \\

SDE$^\dagger$~\cite{kim2023improving} & CVPR'23 & 
    & 80.6 & 96.3 & 98.8 & 64.7 & 91.4 & 96.2 & 528.0
    & 60.4 & 86.2 & 92.4 & 42.6 & 73.1 & 83.1 & 437.8 \\

HREM$^\dagger$~\cite{fu2023learning} & CVPR'23 & 
    & 81.2 & 96.5 & 98.9 & 63.7 & 90.7 & 96.0 & 527.0
    & 60.6 & 86.4 & 92.5 & 41.3 & 71.9 & 82.4 & 435.1 \\

\rowcolor{Gray}
\textbf{Ours} & & & 80.9 & 96.3 & 98.8 & 64.9 & 91.3 & 96.4 & \underline{528.5}
    & 61.4 & 85.6 & 92.4 & 43.3 & 72.9 & 83.3 & \underline{438.9} \\
\rowcolor{Gray}
\textbf{Ours}$^\dagger$ & & & 81.7 & 96.7 & 99.0 & 66.0 & 92.0 & 96.7 & \textbf{532.1}
    & 63.0 & 86.8 & 92.7 & 44.2 & 73.9 & 84.0 & \textbf{444.6} \\

\midrule %
\multicolumn{16}{l}{\textit{\textbf{Faster R-CNN + BERT}}} \\ \midrule

VSE$_\infty$~\cite{chen2021learning} & CVPR'21 &
    & 79.7 & 96.4 & 98.9 & 64.8 & 91.4 & 96.3 & 527.5 
    & 58.3 & 85.3 & 92.3 & 42.4 & 72.7 & 83.2 & 434.2 \\

CODER~\cite{wang2022coder} & ECCV'22 & \cmark
    & 82.1 & 96.6 & 98.8 & 65.5 & 91.5 & 96.2 & 530.7
    & 62.6 & 86.6 & 93.1 & 42.5 & 73.1 & 83.3 & 441.2 \\

MV-VSE$^\dagger$~\cite{li2022multi} & IJCAI'22 & 
    & 80.4 & 96.6 & 99.0 & 64.9 & 91.2 & 96.0 & 528.1
    & 59.1 & 86.3 & 92.5 & 42.5 & 72.8 & 83.1 & 436.3 \\

CHAN~\cite{pan2023fine} & CVPR'23 & \cmark 
    & 81.4 & 96.9 & 98.9 & 66.5 & 92.1 & 96.7 & 532.5
    & 59.8 & 87.2 & 93.3 & 44.9 & 74.5 & 84.2 & 443.9 \\

HREM$^\dagger$~\cite{fu2023learning} & CVPR'23 & 
    & 82.9 & 96.9 & 99.0 & 67.1 & 92.0 & 96.6 & \underline{534.5}
    & 64.0 & 88.5 & 93.7 & 45.4 & 75.1 & 84.3 & \textbf{451.0} \\

\rowcolor{Gray}
\textbf{Ours} & & & 82.4 & 96.8 & 98.8 & 66.2 & 91.9 & 96.6 & 532.7
    & 62.4 & 86.8 & 92.6 & 44.2 & 73.6 & 83.9 & 443.6 \\
\rowcolor{Gray}
\textbf{Ours}$^\dagger$ & & & 82.8 & 97.3 & 99.0 & 67.3 & 92.4 & 96.9 & \textbf{535.6}
    & 64.3 & 87.5 & 93.6 & 45.4 & 74.7 & 84.6 & \underline{450.1} \\

\bottomrule %
\end{tabular}
}
\vspace{-3mm}

\label{tab:coco_comparison} 
\end{table*}

We summarize our results compared with SOTA methods on Flickr30K and MS-COCO in~\cref{tab:f30k_comparison} and~\cref{tab:coco_comparison}. The methods are denoted with whether they are cross-attention or dual-encoder approaches, and are divided into groups depending on the textual backbone used (Bi-GRU vs. BERT). Following previous work~\cite{zhang2022negative,kim2023improving,fu2023learning}, we also report the ensemble results which are obtained by averaging the similarities from two checkpoints trained with different seeds. 

\mypara{Comparisons with state-of-the-art methods} 
When using Bi-GRU as the semantic concept encoder, our method CORA outperforms all state-of-the-art methods by an impressive margin. CORA achieves $+5.4$ RSUM absolute improvement over HREM on Flickr30K, and $+13.7$ RSUM over NAAF on MS-COCO 5K. Note that NAAF is among the SOTA cross-attention methods (CHAN, GraDual, CODER, SGARF) which are more computationally expensive but having more learning capacity advantage over dual encoders, however CORA is still able to surpass them. The non-ensemble version of CORA also outperforms all non-ensemble methods while even exceeding the ensemble ones (SDE, GraDual).

When using BERT for encoding semantic concepts, CORA achieves second best RSUM score on Flickr30K and MS-COCO and is only inferior to the recent SOTA HREM. HREM is also a dual encoder, but is trained with a cross-modality mechanism (which is later discarded at inference) to enhance each modality embedding for matching. The same idea of HREM can be applied to CORA to boost the performance even further, but is out of the scope of our work. Switching from Bi-GRU to using BERT, our method enjoys a smaller RSUM improvement compared to other work ($+5.5$ RSUM for CORA vs. $+15.9$ for HREM and $+10.5$ for CHAN on MS-COCO 5K). This is due to BERT being more suitable for encoding long text while CORA is using BERT to encode short phrases (\eg, \textit{construction worker, sitting}). A text encoder that is more suitable for encoding short phrases is therefore more desirable and we leave this as future work. Similarly, when using BERT, CORA surpasses the performance of SOTA cross-attention methods CODER and CHAN. This shows the ability to generalize across different feature extractors of CORA.

\mypara{Comparisons with scene graph-based approaches} CORA outperforms all scene-graph based methods, which includes SGM~\cite{wang2020cross}, GCN+DIST~\cite{li2020visual}, GSMN~\cite{liu2020graph}, and GraDual~\cite{long2022gradual}. These methods all employ an additional off-the-shelf visual scene graph generator~\cite{zellers2018neural} (except GSMN) to produce a scene graph for an input image, and use cross-attention to exchange information between the textual and the visual graph, but still achieve inferior results to CORA. This further shows that CORA is very effective at encoding scene graphs. These methods all embed a whole scene graph holistically (unlike CORA which separates it into two object-attribute and object-object steps), are trained with a holistic loss to align image and text (unlike CORA that has loss terms to additionally align image and local object entity), and use visual scene graph generator~\cite{zellers2018neural} which is susceptible to making wrong predictions and has been reported to misdetect rare object relationships~\cite{tang2020unbiased}.

\subsection{Ablation Studies}
\label{sec:ablation}
We perform a series of ablation studies to explore the impact of our graph attention network design and how the losses affect the final performance. All experiments in this section use Bi-GRU for the semantic encoder and are performed on the Flickr30K dataset. The results are reported in~\cref{tab:ablation_study}.

\begin{table}[t]
\centering
\caption{
\textbf{Ablation studies} for the number of layers in GAT, the graph structure whether encoding scene graph jointly or in 2 separate steps is beneficial, and the impact of losses. \textbf{Bold} and \underline{underline} highlight the best and second-best performance.
}
\scalebox{0.7}{
    \setlength{\tabcolsep}{3.5pt}
    \begin{tabular}{@{}p{0.01cm}p{0.01cm}p{0.01cm}p{0.01cm}p{0.01cm}p{0.01cm}|ccc|ccc|c@{}}
    \toprule
    
    & & & & & 
    & \multicolumn{3}{c|}{Image $\rightarrow$ Text} 
    & \multicolumn{3}{c|}{Text $\rightarrow$ Image} 
    & \multirow{2}{*}{RSUM} \\ \cmidrule{7-12}
    
    & & & & & & R@1 & R@5 & R@10 & R@1 & R@5 & R@10 & \\ \midrule
    
    \multicolumn{6}{c}{\textit{\textbf{Number of GAT layers}}} \\ 
    \multicolumn{3}{c}{$n_\text{Obj-Att}$} & \multicolumn{3}{c}{$n_\text{Obj-Obj}$} \\ \midrule
    \multicolumn{3}{c}{1} & \multicolumn{3}{c|}{1} & 79.8 & 95.3 & 97.1 & 60.5 & 85.4 & 91.0 & 509.1 \\
    \multicolumn{3}{c}{1} & \multicolumn{3}{c|}{2} & 82.2 & 95.6 & 97.7 & 61.8 & 86.5 & 92.0 & \textbf{515.8} \\
    \multicolumn{3}{c}{1} & \multicolumn{3}{c|}{3} & 81.6 & 95.8 & 97.6 & 61.3 & 86.2 & 91.9 & 514.4 \\
    \multicolumn{3}{c}{2} & \multicolumn{3}{c|}{1} & 79.9 & 95.1 & 97.5 & 60.8 & 85.5 & 91.3 & 510.1 \\
    \multicolumn{3}{c}{2} & \multicolumn{3}{c|}{2} & 81.7 & 95.3 & 96.8 & 61.6 & 87.0 & 92.1 & 514.5 \\
    \multicolumn{3}{c}{2} & \multicolumn{3}{c|}{3} & 80.2 & 95.3 & 96.5 & 60.9 & 85.9 & 91.8 & 510.6 \\ \midrule

    \multicolumn{6}{c}{\textit{\textbf{Graph structure}}} \\ 
    \multicolumn{2}{c}{Joint}  & \multicolumn{2}{c}{\scriptsize{Obj-Att\&Obj-Obj}} & \multicolumn{2}{c}{FC} \\ \midrule
    \multicolumn{2}{c}{\cmark} & & & &  & 78.9 & 93.5 & 96.3 & 59.6 & 85.8 & 90.4 & 504.8 \\
    \multicolumn{2}{c}{\cmark} & & & \multicolumn{2}{c|}{\cmark} & 77.6 & 93.0 & 96.0 & 59.4 & 85.1 & 90.5 & 501.6 \\
    & & \multicolumn{2}{c}{\cmark} & & & 82.2 & 95.6 & 97.7 & 61.8 & 86.5 & 92.0 & \textbf{515.8} \\
    & & \multicolumn{2}{c}{\cmark} & \multicolumn{2}{c|}{\cmark} & 81.2 & 95.1 & 96.9 & 61.3 & 86.9 & 91.8 & 513.2 \\ \midrule

    \multicolumn{6}{c}{\textit{\textbf{Losses}}} \\
    \multicolumn{2}{c}{$\mathcal{L}_\text{HARD}$}  & \multicolumn{2}{c}{$\mathcal{L}_\text{CON}$} & \multicolumn{2}{c}{$\mathcal{L}_\text{SPEC}$} \\ \midrule
    \multicolumn{2}{c}{\cmark} & & & &  & 71.6 & 92.2 & 95.9 & 53.9 & 83.4 & 89.9 & 486.9 \\
     & & \multicolumn{2}{c}{\cmark} & & & 75.0 & 94.0 & 96.4 & 57.9 & 84.4 & 90.1 & 497.8 \\
    \multicolumn{2}{c}{\cmark} & \multicolumn{2}{c}{\cmark} & & & 78.9 & 95.2 & 97.3 & 59.1 & 85.8 & 91.0 & 507.3 \\
    \multicolumn{2}{c}{\cmark} & \multicolumn{2}{c}{\cmark} & \multicolumn{2}{c|}{\cmark} & 82.2 & 95.6 & 97.7 & 61.8 & 86.5 & 92.0 & \textbf{515.8} \\
    \bottomrule
    \end{tabular}
}

\vspace{-3mm}
\label{tab:ablation_study}
\end{table}

\mypara{Number of layers in GAT} The experiments show that having $1$ layer for $\text{GAT}_\text{Obj-Att}$ and 2 layers for $\text{GAT}_\text{Obj-Obj}$ achieves the best accuracy. For the object-attribute graph, 1 layer is sufficient to propagate the attribute information to their corresponding object node. For the object-object relation graph, using only 1 layer is not enough to aggregate information from the whole graph, while increasing to 3 layers starts to give diminishing returns.

\mypara{Graph structure} We study whether our 2-step scene graph encoding step is beneficial to the final performance. We refer to \textit{Joint} as the model that uses a single GAT on the whole graph at once, \textit{FC} as the variant that uses fully connected graph instead of the structure parsed from the scene graph parser, and \textit{Obj-Att \& Obj-Obj} as our proposed 2-step scene graph encoding model. Note that having a separate object-attribute encoding step allows our model to produce individual entity embeddings (see~\cref{subsec:objatt_gat}) that are later used in the contrastive loss to align an image with each of its objects independently. This is not possible with the \textit{Joint} model because after the GAT step, all object nodes are already contextualized. The results indeed show that our proposed 2-step encoding step is superior, and using the graph structure parsed from the scene graph parser is better than connecting all nodes together in a fully connected manner.

\mypara{Losses} Similar to prior work, we first explore using the triplet loss with hardest negatives $\mathcal{L}_\text{HARD}$ and find that using it alone is insufficient, since it can only be used on the image-text level and not at the image-object entity level. Using solely the contrastive loss $\mathcal{L}_\text{CON}$ gives an accuracy boost, but is still far from optimal. By combining $\mathcal{L}_\text{HARD}$ and $\mathcal{L}_\text{CON}$, we achieve a significant accuracy improvement. Since the contrastive loss treats the graph embedding and entity embedding equally, by imposing the structure that a whole text caption should depict more information than an entity alone through the loss $\mathcal{L}_\text{SPEC}$, CORA achieves SOTA results.

\subsection{Qualitative Results \& Text-to-Entity Retrieval}
\begin{figure}
\centering
\includegraphics[width=1.0\linewidth]{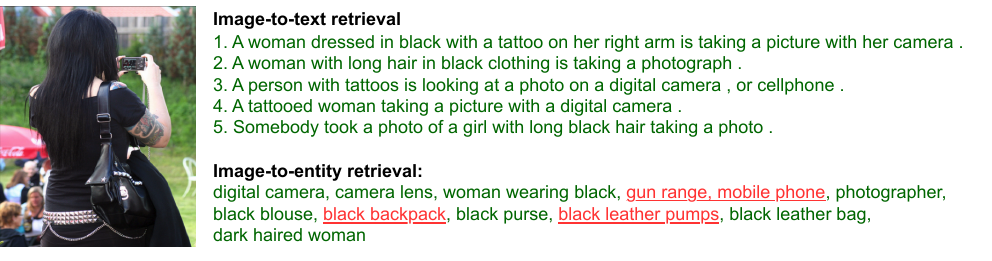}
    \caption{\textbf{Qualitative result} demonstrates how CORA can perform image-to-text and image-to-entity retrieval. {\color{greentext}Green} denotes correct retrieval while {\color{red} \underline{red}} denotes incorrect ones.}
    \vspace{-3mm}
\label{fig:qualitative}
\end{figure}

\cref{fig:qualitative} illustrates how CORA can perform image-to-text and image-to-object entity retrieval. More qualitative results can be found in the supplementary.

In addition, we also experiment with using the image-entity score for re-ranking the image-text matching. We employ the idea that if an object entity in the text is not closely matched with the image, then the image-text matching score should be lower. Formally, we utilize the following formula
\begin{align}
    \hat{s}(v_i, t_i) = \beta \cdot s(v_i, t_i) + (1 - \beta) \cdot \min_k s(v_i, e_{ik}),
\end{align}
where $\beta \in [0, 1]$ is a hyperparameter that we select on the validation set. The results on MS-COCO are reported in~\cref{tab:reranking_results}, where we achieve a slight accuracy improvement with this simple strategy. One potential future direction is to explore smarter mechanism to combine the image-text and image-object entity embedding alignment score.

\begin{table}[t]
\centering
\caption{
\textbf{Reranking results on MS-COCO 5K} after ensembling with image-entity score.
}
\setlength{\tabcolsep}{5.pt}
\scalebox{0.7}{
    \begin{tabular}{@{}l|ccc|ccc|c@{}}
    \toprule
    
    \multirow{2}{*}{}
    & \multicolumn{3}{c|}{Image $\rightarrow$ Text} 
    & \multicolumn{3}{c|}{Text $\rightarrow$ Image} 
    & \multirow{2}{*}{RSUM} \\
    \cmidrule{2-7}
    & R@1 & R@5 & R@10 & R@1 & R@5 & R@10 & \\ \midrule
    
    \multicolumn{8}{l}{\textit{\textbf{Faster R-CNN + Bi-GRU}}} \\ \midrule

    CORA & 63.0 & 86.8 & 92.7 & 44.2 & 73.9 & 84.0 & 444.6 \\

    CORA + reranking & 63.2 & 86.8 & 92.7 & 44.3 & 74.1 & 84.0 & 445.1 \\

    \midrule

    \multicolumn{8}{l}{\textit{\textbf{Faster R-CNN + BERT}}} \\ \midrule

    CORA & 64.3 & 87.5 & 93.6 & 45.4 & 74.7 & 84.6 & 450.1 \\
    
    CORA + reranking & 64.2 & 87.6 & 93.8 & 45.5 & 74.8 & 84.7 & 450.6 \\

    \bottomrule
    \end{tabular}
}

\label{tab:reranking_results}
\end{table}
 
\section{Conclusion}
\label{sec:conclusion}
\mypara{Limitation} Despite achieving new SOTA results, CORA still faces some limitations. CORA is strongly dependent on the scene graph quality from the parser. If the parser fails to extract a scene graph from the input text, CORA also fails to encode the text. This happens seldomly in MS-COCO, where there are captions that are just exclamatory sentences uttered by the annotator, \eg, ``I am so happy to see this view'', ``There are so many things to see here.'' On the other hand, text sequence model is still able to capture the nuances of these text descriptions.

In this paper, we propose a dual-encoder model CORA for image-text matching that is based on scene graph. CORA achieves new SOTA results, outperforms all SOTA computationally expensive cross-attention methods. We show a promising future direction for image-text matching that, by representing a caption as a scene graph of object and attribute nodes connected by relation edges, we can utilize the strong relational inductive bias of graph neural network to compose objects, relations, and their attributes into a scene graph embedding that is effective for image-text retrieval.

\smallskip
\noindent\textbf{Acknowledgements.} This project was partially funded by NSF CAREER Award (\#2238769) to AS.
{
    \small
    \bibliographystyle{ieeenat_fullname}
    \bibliography{main}

\begin{thebibliography}{58}
\providecommand{\natexlab}[1]{#1}
\providecommand{\url}[1]{\texttt{#1}}
\expandafter\ifx\csname urlstyle\endcsname\relax
  \providecommand{\doi}[1]{doi: #1}\else
  \providecommand{\doi}{doi: \begingroup \urlstyle{rm}\Url}\fi

\bibitem[Anderson et~al.(2018)Anderson, He, Buehler, Teney, Johnson, Gould, and Zhang]{anderson2018bottom}
Peter Anderson, Xiaodong He, Chris Buehler, Damien Teney, Mark Johnson, Stephen Gould, and Lei Zhang.
\newblock Bottom-up and top-down attention for image captioning and visual question answering.
\newblock In \emph{CVPR}, 2018.

\bibitem[Brody et~al.(2021)Brody, Alon, and Yahav]{brody2021attentive}
Shaked Brody, Uri Alon, and Eran Yahav.
\newblock How attentive are graph attention networks?
\newblock \emph{arXiv preprint arXiv:2105.14491}, 2021.

\bibitem[Chefer et~al.(2023)Chefer, Alaluf, Vinker, Wolf, and Cohen-Or]{chefer2023attend}
Hila Chefer, Yuval Alaluf, Yael Vinker, Lior Wolf, and Daniel Cohen-Or.
\newblock Attend-and-excite: Attention-based semantic guidance for text-to-image diffusion models.
\newblock \emph{arXiv preprint arXiv:2301.13826}, 2023.

\bibitem[Chen et~al.(2021)Chen, Hu, Wu, Jiang, and Wang]{chen2021learning}
Jiacheng Chen, Hexiang Hu, Hao Wu, Yuning Jiang, and Changhu Wang.
\newblock Learning the best pooling strategy for visual semantic embedding.
\newblock In \emph{CVPR}, 2021.

\bibitem[Chen et~al.(2020)Chen, Li, Yu, El~Kholy, Ahmed, Gan, Cheng, and Liu]{chen2020uniter}
Yen-Chun Chen, Linjie Li, Licheng Yu, Ahmed El~Kholy, Faisal Ahmed, Zhe Gan, Yu Cheng, and Jingjing Liu.
\newblock Uniter: Universal image-text representation learning.
\newblock In \emph{ECCV}. Springer, 2020.

\bibitem[Chun et~al.(2021)Chun, Oh, De~Rezende, Kalantidis, and Larlus]{chun2021probabilistic}
Sanghyuk Chun, Seong~Joon Oh, Rafael~Sampaio De~Rezende, Yannis Kalantidis, and Diane Larlus.
\newblock Probabilistic embeddings for cross-modal retrieval.
\newblock In \emph{CVPR}, 2021.

\bibitem[Chung et~al.(2014)Chung, Gulcehre, Cho, and Bengio]{chung2014empirical}
Junyoung Chung, Caglar Gulcehre, KyungHyun Cho, and Yoshua Bengio.
\newblock Empirical evaluation of gated recurrent neural networks on sequence modeling.
\newblock \emph{arXiv preprint arXiv:1412.3555}, 2014.

\bibitem[Devlin et~al.(2018)Devlin, Chang, Lee, and Toutanova]{devlin2018bert}
Jacob Devlin, Ming-Wei Chang, Kenton Lee, and Kristina Toutanova.
\newblock Bert: Pre-training of deep bidirectional transformers for language understanding.
\newblock \emph{arXiv preprint arXiv:1810.04805}, 2018.

\bibitem[Diao et~al.(2021)Diao, Zhang, Ma, and Lu]{diao2021similarity}
Haiwen Diao, Ying Zhang, Lin Ma, and Huchuan Lu.
\newblock Similarity reasoning and filtration for image-text matching.
\newblock In \emph{AAAI}, 2021.

\bibitem[Faghri et~al.(2017)Faghri, Fleet, Kiros, and Fidler]{faghri2017vse++}
Fartash Faghri, David~J Fleet, Jamie~Ryan Kiros, and Sanja Fidler.
\newblock Vse++: Improving visual-semantic embeddings with hard negatives.
\newblock \emph{arXiv preprint arXiv:1707.05612}, 2017.

\bibitem[Frome et~al.(2013)Frome, Corrado, Shlens, Bengio, Dean, Ranzato, and Mikolov]{frome2013devise}
Andrea Frome, Greg~S Corrado, Jon Shlens, Samy Bengio, Jeff Dean, Marc'Aurelio Ranzato, and Tomas Mikolov.
\newblock Devise: A deep visual-semantic embedding model.
\newblock \emph{NeurIPS}, 26, 2013.

\bibitem[Fu et~al.(2023)Fu, Mao, Song, and Zhang]{fu2023learning}
Zheren Fu, Zhendong Mao, Yan Song, and Yongdong Zhang.
\newblock Learning semantic relationship among instances for image-text matching.
\newblock In \emph{CVPR}, 2023.

\bibitem[He et~al.(2016)He, Zhang, Ren, and Sun]{he2016deep}
Kaiming He, Xiangyu Zhang, Shaoqing Ren, and Jian Sun.
\newblock Deep residual learning for image recognition.
\newblock In \emph{CVPR}, 2016.

\bibitem[Hewitt and Manning(2019)]{hewitt2019structural}
John Hewitt and Christopher~D Manning.
\newblock A structural probe for finding syntax in word representations.
\newblock In \emph{Proceedings of the 2019 Conference of the North American Chapter of the Association for Computational Linguistics: Human Language Technologies, Volume 1 (Long and Short Papers)}, pages 4129--4138, 2019.

\bibitem[Hochreiter and Schmidhuber(1997)]{hochreiter1997long}
Sepp Hochreiter and J{\"u}rgen Schmidhuber.
\newblock Long short-term memory.
\newblock \emph{Neural computation}, 9\penalty0 (8):\penalty0 1735--1780, 1997.

\bibitem[Honnibal et~al.(2020)Honnibal, Montani, Van~Landeghem, and Boyd]{spacy}
Matthew Honnibal, Ines Montani, Sofie Van~Landeghem, and Adriane Boyd.
\newblock {spaCy: Industrial-strength Natural Language Processing in Python}.
\newblock 2020.

\bibitem[Huang et~al.(2019)Huang, Liang, Duan, Gong, Shou, Jiang, and Zhou]{huang2019unicoder}
Haoyang Huang, Yaobo Liang, Nan Duan, Ming Gong, Linjun Shou, Daxin Jiang, and Ming Zhou.
\newblock Unicoder: A universal language encoder by pre-training with multiple cross-lingual tasks.
\newblock \emph{arXiv preprint arXiv:1909.00964}, 2019.

\bibitem[Huynh et~al.(2023)Huynh, Zhou, Lin, Barnes, Shechtman, Amirghodsi, and Shrivastava]{huynh2023simpson}
Chuong Huynh, Yuqian Zhou, Zhe Lin, Connelly Barnes, Eli Shechtman, Sohrab Amirghodsi, and Abhinav Shrivastava.
\newblock Simpson: Simplifying photo cleanup with single-click distracting object segmentation network.
\newblock In \emph{CVPR}, 2023.

\bibitem[Huynh et~al.(2024)Huynh, Oh, , Shrivastava, and Lee]{huynh2024maggie}
Chuong Huynh, Seoung~Wug Oh, , Abhinav Shrivastava, and Joon-Young Lee.
\newblock Maggie: Masked guided gradual human instance matting.
\newblock In \emph{CVPR}, 2024.

\bibitem[Kim et~al.(2023)Kim, Kim, and Kwak]{kim2023improving}
Dongwon Kim, Namyup Kim, and Suha Kwak.
\newblock Improving cross-modal retrieval with set of diverse embeddings.
\newblock In \emph{CVPR}, 2023.

\bibitem[Kiros et~al.(2014)Kiros, Salakhutdinov, and Zemel]{kiros2014unifying}
Ryan Kiros, Ruslan Salakhutdinov, and Richard~S Zemel.
\newblock Unifying visual-semantic embeddings with multimodal neural language models.
\newblock \emph{arXiv preprint arXiv:1411.2539}, 2014.

\bibitem[Krishna et~al.(2017)Krishna, Zhu, Groth, Johnson, Hata, Kravitz, Chen, Kalantidis, Li, Shamma, et~al.]{krishna2017visual}
Ranjay Krishna, Yuke Zhu, Oliver Groth, Justin Johnson, Kenji Hata, Joshua Kravitz, Stephanie Chen, Yannis Kalantidis, Li-Jia Li, David~A Shamma, et~al.
\newblock Visual genome: Connecting language and vision using crowdsourced dense image annotations.
\newblock \emph{IJCV}, 2017.

\bibitem[Lee et~al.(2018)Lee, Chen, Hua, Hu, and He]{lee2018stacked}
Kuang-Huei Lee, Xi Chen, Gang Hua, Houdong Hu, and Xiaodong He.
\newblock Stacked cross attention for image-text matching.
\newblock In \emph{ECCV}, 2018.

\bibitem[Li et~al.(2019)Li, Zhang, Li, Li, and Fu]{li2019visual}
Kunpeng Li, Yulun Zhang, Kai Li, Yuanyuan Li, and Yun Fu.
\newblock Visual semantic reasoning for image-text matching.
\newblock In \emph{ICCV}, pages 4654--4662, 2019.

\bibitem[Li et~al.(2020)Li, Zhang, and Mu]{li2020visual}
Yongzhi Li, Duo Zhang, and Yadong Mu.
\newblock Visual-semantic matching by exploring high-order attention and distraction.
\newblock In \emph{CVPR}, 2020.

\bibitem[Li et~al.(2022)Li, Guo, Feng, Hwang, and Xue]{li2022multi}
Zheng Li, Caili Guo, Zerun Feng, Jenq-Neng Hwang, and Xijun Xue.
\newblock Multi-view visual semantic embedding.
\newblock In \emph{IJCAI}, page~7, 2022.

\bibitem[Lin et~al.(2014)Lin, Maire, Belongie, Hays, Perona, Ramanan, Doll{\'a}r, and Zitnick]{lin2014microsoft}
Tsung-Yi Lin, Michael Maire, Serge Belongie, James Hays, Pietro Perona, Deva Ramanan, Piotr Doll{\'a}r, and C~Lawrence Zitnick.
\newblock Microsoft coco: Common objects in context.
\newblock In \emph{ECCV}, 2014.

\bibitem[Liu et~al.(2020)Liu, Mao, Zhang, Xie, Wang, and Zhang]{liu2020graph}
Chunxiao Liu, Zhendong Mao, Tianzhu Zhang, Hongtao Xie, Bin Wang, and Yongdong Zhang.
\newblock Graph structured network for image-text matching.
\newblock In \emph{CVPR}, 2020.

\bibitem[Liu et~al.(2021)Liu, Ji, Fu, Tam, Du, Yang, and Tang]{liu2021p}
Xiao Liu, Kaixuan Ji, Yicheng Fu, Weng~Lam Tam, Zhengxiao Du, Zhilin Yang, and Jie Tang.
\newblock P-tuning v2: Prompt tuning can be comparable to fine-tuning universally across scales and tasks.
\newblock \emph{arXiv preprint arXiv:2110.07602}, 2021.

\bibitem[Long et~al.(2022)Long, Han, Wan, and Poon]{long2022gradual}
Siqu Long, Soyeon~Caren Han, Xiaojun Wan, and Josiah Poon.
\newblock Gradual: Graph-based dual-modal representation for image-text matching.
\newblock In \emph{WACV}, 2022.

\bibitem[Loshchilov and Hutter(2017)]{loshchilov2017decoupled}
Ilya Loshchilov and Frank Hutter.
\newblock Decoupled weight decay regularization.
\newblock \emph{arXiv preprint arXiv:1711.05101}, 2017.

\bibitem[Lu et~al.(2019)Lu, Batra, Parikh, and Lee]{lu2019vilbert}
Jiasen Lu, Dhruv Batra, Devi Parikh, and Stefan Lee.
\newblock Vilbert: Pretraining task-agnostic visiolinguistic representations for vision-and-language tasks.
\newblock \emph{NeurIPS}, 32, 2019.

\bibitem[Mahajan et~al.(2018)Mahajan, Girshick, Ramanathan, He, Paluri, Li, Bharambe, and Van Der~Maaten]{mahajan2018exploring}
Dhruv Mahajan, Ross Girshick, Vignesh Ramanathan, Kaiming He, Manohar Paluri, Yixuan Li, Ashwin Bharambe, and Laurens Van Der~Maaten.
\newblock Exploring the limits of weakly supervised pretraining.
\newblock In \emph{ECCV}, pages 181--196, 2018.

\bibitem[Nguyen et~al.(2021)Nguyen, Tripathi, Du, Guha, and Nguyen]{nguyen2021defense}
Kien Nguyen, Subarna Tripathi, Bang Du, Tanaya Guha, and Truong~Q Nguyen.
\newblock In defense of scene graphs for image captioning.
\newblock In \emph{ICCV}, 2021.

\bibitem[Pan et~al.(2023)Pan, Wu, and Zhang]{pan2023fine}
Zhengxin Pan, Fangyu Wu, and Bailing Zhang.
\newblock Fine-grained image-text matching by cross-modal hard aligning network.
\newblock In \emph{CVPR}, 2023.

\bibitem[Patterson and Hays(2016)]{patterson2016coco}
Genevieve Patterson and James Hays.
\newblock Coco attributes: Attributes for people, animals, and objects.
\newblock In \emph{ECCV}, pages 85--100. Springer, 2016.

\bibitem[Pennington et~al.(2014)Pennington, Socher, and Manning]{pennington2014glove}
Jeffrey Pennington, Richard Socher, and Christopher~D Manning.
\newblock Glove: Global vectors for word representation.
\newblock In \emph{Proceedings of the 2014 conference on empirical methods in natural language processing (EMNLP)}, 2014.

\bibitem[Pham et~al.(2021)Pham, Kafle, Lin, Ding, Cohen, Tran, and Shrivastava]{pham2021learning}
Khoi Pham, Kushal Kafle, Zhe Lin, Zhihong Ding, Scott Cohen, Quan Tran, and Abhinav Shrivastava.
\newblock Learning to predict visual attributes in the wild.
\newblock In \emph{CVPR}, pages 13018--13028, 2021.

\bibitem[Pham et~al.(2022)Pham, Kafle, Lin, Ding, Cohen, Tran, and Shrivastava]{pham2022improving}
Khoi Pham, Kushal Kafle, Zhe Lin, Zhihong Ding, Scott Cohen, Quan Tran, and Abhinav Shrivastava.
\newblock Improving closed and open-vocabulary attribute prediction using transformers.
\newblock In \emph{ECCV}, 2022.

\bibitem[Phung et~al.(2024)Phung, Ge, and Huang]{phung2024attenref}
Quynh Phung, Songwei Ge, and Jia-Bin Huang.
\newblock Grounded text-to-image synthesis with attention refocusing.
\newblock In \emph{CVPR}, 2024.

\bibitem[Plummer et~al.(2015)Plummer, Wang, Cervantes, Caicedo, Hockenmaier, and Lazebnik]{plummer2015flickr30k}
Bryan~A Plummer, Liwei Wang, Chris~M Cervantes, Juan~C Caicedo, Julia Hockenmaier, and Svetlana Lazebnik.
\newblock Flickr30k entities: Collecting region-to-phrase correspondences for richer image-to-sentence models.
\newblock In \emph{ICCV}, 2015.

\bibitem[Radford et~al.(2021)Radford, Kim, Hallacy, Ramesh, Goh, Agarwal, Sastry, Askell, Mishkin, Clark, et~al.]{radford2021learning}
Alec Radford, Jong~Wook Kim, Chris Hallacy, Aditya Ramesh, Gabriel Goh, Sandhini Agarwal, Girish Sastry, Amanda Askell, Pamela Mishkin, Jack Clark, et~al.
\newblock Learning transferable visual models from natural language supervision.
\newblock In \emph{International conference on machine learning}, pages 8748--8763. PMLR, 2021.

\bibitem[Rombach et~al.(2022)Rombach, Blattmann, Lorenz, Esser, and Ommer]{rombach2022high}
Robin Rombach, Andreas Blattmann, Dominik Lorenz, Patrick Esser, and Bj{\"o}rn Ommer.
\newblock High-resolution image synthesis with latent diffusion models.
\newblock In \emph{CVPR}, 2022.

\bibitem[Saini et~al.(2022)Saini, Pham, and Shrivastava]{saini2022disentangling}
Nirat Saini, Khoi Pham, and Abhinav Shrivastava.
\newblock Disentangling visual embeddings for attributes and objects.
\newblock In \emph{CVPR}, pages 13658--13667, 2022.

\bibitem[Schuster et~al.(2015)Schuster, Krishna, Chang, Fei-Fei, and Manning]{schuster2015generating}
Sebastian Schuster, Ranjay Krishna, Angel Chang, Li Fei-Fei, and Christopher~D. Manning.
\newblock Generating semantically precise scene graphs from textual descriptions for improved image retrieval.
\newblock In \emph{Workshop on Vision and Language (VL15)}, Lisbon, Portugal, 2015. Association for Computational Linguistics.

\bibitem[Song and Soleymani(2019)]{song2019polysemous}
Yale Song and Mohammad Soleymani.
\newblock Polysemous visual-semantic embedding for cross-modal retrieval.
\newblock In \emph{CVPR}, pages 1979--1988, 2019.

\bibitem[Tang et~al.(2020)Tang, Niu, Huang, Shi, and Zhang]{tang2020unbiased}
Kaihua Tang, Yulei Niu, Jianqiang Huang, Jiaxin Shi, and Hanwang Zhang.
\newblock Unbiased scene graph generation from biased training.
\newblock In \emph{CVPR}, 2020.

\bibitem[Veli{\v{c}}kovi{\'c} et~al.(2017)Veli{\v{c}}kovi{\'c}, Cucurull, Casanova, Romero, Lio, and Bengio]{velivckovic2017graph}
Petar Veli{\v{c}}kovi{\'c}, Guillem Cucurull, Arantxa Casanova, Adriana Romero, Pietro Lio, and Yoshua Bengio.
\newblock Graph attention networks.
\newblock \emph{arXiv preprint arXiv:1710.10903}, 2017.

\bibitem[Wang et~al.(2022)Wang, He, Wu, Xia, Yang, Li, Yu, Ji, Ding, and Wang]{wang2022coder}
Haoran Wang, Dongliang He, Wenhao Wu, Boyang Xia, Min Yang, Fu Li, Yunlong Yu, Zhong Ji, Errui Ding, and Jingdong Wang.
\newblock Coder: Coupled diversity-sensitive momentum contrastive learning for image-text retrieval.
\newblock In \emph{ECCV}, 2022.

\bibitem[Wang et~al.(2016)Wang, Li, and Lazebnik]{wang2016learning}
Liwei Wang, Yin Li, and Svetlana Lazebnik.
\newblock Learning deep structure-preserving image-text embeddings.
\newblock In \emph{CVPR}, 2016.

\bibitem[Wang et~al.(2020)Wang, Wang, Yao, Shan, and Chen]{wang2020cross}
Sijin Wang, Ruiping Wang, Ziwei Yao, Shiguang Shan, and Xilin Chen.
\newblock Cross-modal scene graph matching for relationship-aware image-text retrieval.
\newblock In \emph{WACV}, 2020.

\bibitem[Wen et~al.(2020)Wen, Gu, and Cheng]{wen2020learning}
Keyu Wen, Xiaodong Gu, and Qingrong Cheng.
\newblock Learning dual semantic relations with graph attention for image-text matching.
\newblock \emph{IEEE transactions on circuits and systems for video technology}, 31\penalty0 (7):\penalty0 2866--2879, 2020.

\bibitem[Wu et~al.(2019{\natexlab{a}})Wu, Mao, Zhang, Jiang, Li, Sun, and Ma]{wu2019unified}
Hao Wu, Jiayuan Mao, Yufeng Zhang, Yuning Jiang, Lei Li, Weiwei Sun, and Wei-Ying Ma.
\newblock Unified visual-semantic embeddings: Bridging vision and language with structured meaning representations.
\newblock In \emph{CVPR}, 2019{\natexlab{a}}.

\bibitem[Wu et~al.(2019{\natexlab{b}})Wu, Wang, Song, and Huang]{wu2019learning}
Yiling Wu, Shuhui Wang, Guoli Song, and Qingming Huang.
\newblock Learning fragment self-attention embeddings for image-text matching.
\newblock In \emph{ACM MM}, 2019{\natexlab{b}}.

\bibitem[Xie et~al.(2017)Xie, Girshick, Doll{\'a}r, Tu, and He]{xie2017aggregated}
Saining Xie, Ross Girshick, Piotr Doll{\'a}r, Zhuowen Tu, and Kaiming He.
\newblock Aggregated residual transformations for deep neural networks.
\newblock In \emph{CVPR}, pages 1492--1500, 2017.

\bibitem[Zellers et~al.(2018)Zellers, Yatskar, Thomson, and Choi]{zellers2018neural}
Rowan Zellers, Mark Yatskar, Sam Thomson, and Yejin Choi.
\newblock Neural motifs: Scene graph parsing with global context.
\newblock In \emph{CVPR}, 2018.

\bibitem[Zhang et~al.(2022)Zhang, Mao, Wang, and Zhang]{zhang2022negative}
Kun Zhang, Zhendong Mao, Quan Wang, and Yongdong Zhang.
\newblock Negative-aware attention framework for image-text matching.
\newblock In \emph{CVPR}, 2022.

\bibitem[Zhang et~al.(2020)Zhang, Lei, Zhang, and Li]{zhang2020context}
Qi Zhang, Zhen Lei, Zhaoxiang Zhang, and Stan~Z Li.
\newblock Context-aware attention network for image-text retrieval.
\newblock In \emph{CVPR}, 2020.

\end{thebibliography}
}

\clearpage
\setcounter{page}{1}
\maketitlesupplementary

\section{Implementation Details}
\label{sec:implementation}
\subsection{Visual features}
We use the pre-extracted Faster R-CNN 2048-dimensional region features from BUTD~\cite{lee2018stacked,anderson2018bottom}, which is the standard convention from prior work in the image-text matching literature~\cite{chen2021learning,fu2023learning}. To transform them to have the same dimensions $D$ as the joint embedding space, we implement a 2-layer MLP with residual connection. The region features are then pooled using the GPO~\cite{chen2021learning} pooling operator into $\mathbb{R}^D$.

\subsection{Textual features}
\mypara{Bi-GRU} The dimension of the word embedding is set to 300 for both experiments where we initialize the word embedding from GloVe or from scratch (refer to~\cref{sec:more_ablation} for experiment of CORA without using GloVe). The GRU has 1 layer and its hidden dimension is also 300. 

\mypara{BERT}
Similar to prior work, we use the \texttt{bert-base-uncased} architecture and pre-trained weights for the BERT semantic concept encoder. As mentioned in the main paper, using BERT to encode short phrases (\eg, \textit{construction worker, sitting}) does not take advantage of the full capability of BERT. BERT has never seen short text during its pre-training stage~\cite{devlin2018bert}, and with its
ability to capture long-range dependencies, BERT is more suitable for encoding long sentences. As a result, direct fine-tuning BERT for CORA leads to slightly lower results.

In our work, instead of fine-tuning the whole BERT model (with 110M params), we employ the prefix tuning technique P-Tuning v2~\cite{liu2021p} in order to repurpose the pre-trained BERT model into encoding short phrases. With this technique, at every BERT encoding layer, a sequence of learnable $N$ token embeddings $\mathbb{R}^{N \times 768}$ is added as prefix into the textual prompt. Intuitively, these tokens provide learnable context that assist BERT into learning the task at hand, which is encoding short phrases. The number of trainable params with P-Tuning v2 is only $2N \times L \times 768$ (where $L=12$ is the number of BERT encoding layers, and $N=24$ is the number of prefix tokens). In our experiment, we find fine-tuning the last BERT layer along with P-Tuning gives slightly better results. In overall, the number of trainable params of our BERT component is only 7M, which is much smaller than 110M params of the whole BERT model.


For both types of features (Bi-GRU and BERT), we implement an FC layer to transform the semantic encoded output into $\mathbb{R}^D$ before using them to initialize the node and edge features of the GATs.

\subsection{Training and hyperparameter details}
We use the AdamW optimizer~\cite{loshchilov2017decoupled} to train our model for 50 epochs. The learning rate is initialized at 5e-4, then decayed to 5e-5 after 15 epochs. The learning rate for the pre-trained components (\ie, GloVe and BERT) is scaled by 0.1 w.r.t. the base learning rate.
We set the batch size to 128 when training on Flickr30K, and 256 when training on MS-COCO.
The margin $\alpha$ in the triplet loss is set to 0.4, while the cosine similarity in the contrastive loss is scaled by a temperature of 0.01 similar to CLIP~\cite{radford2021learning}. Following~\cite{chen2021learning}, we perform size augmentation to randomly drop 35\% region features. For data augmentation on the text, we perform subsampling on the scene graph by randomly dropping 10\% of the nodes and edges and randomly masking 10\% of the word tokens. We set $\lambda_\text{CON}=0.25$ and $\lambda_\text{SPEC}=3.0$.

\section{Inference Time}
\begin{figure}[t]
\centering
\includegraphics[width=1.0\linewidth]{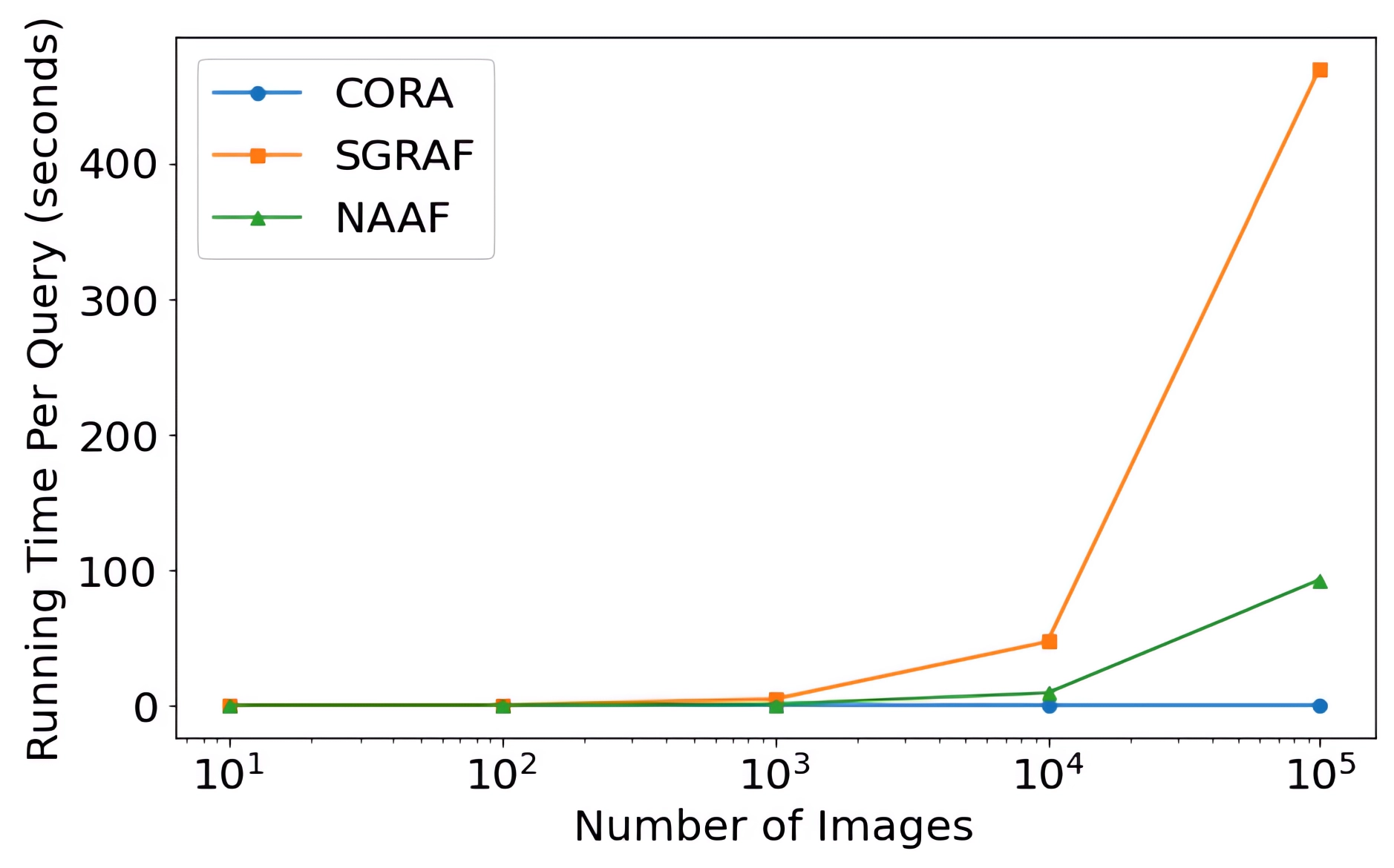}
    \caption{\textbf{Inference time comparison.} We compare the text-to-image retrieval inference time between our method CORA against two SOTA cross-attention methods SGRAF~\cite{diao2021similarity} and NAAF~\cite{zhang2022negative} (lower is better). The inference time is calculated with different number of images in the database. CORA with its dual-encoder architecture is much faster and scalable than cross-attention approaches.}
\label{fig:inference_time}
\vspace{-1mm}
\end{figure}
We illustrate in~\cref{fig:inference_time} the inference time comparison between our method CORA against SOTA cross-attention methods SGRAF~\cite{diao2021similarity} and NAAF~\cite{zhang2022negative} with different number of images in the database (ranging from a very small to a very large number of images).

To conduct this experiment, for all methods, we first forward all images through the image encoder of each respective method in order to cache all image embeddings. Then, for CORA, when a text query arrives, it takes 0.04s to parse it into a scene graph, 0.014s to compute its scene graph embedding, then 0.01s to perform the vector-matrix multiplication with the image embeddings to find nearest neighbor results, which in total accounts to around 0.06s per query for all number of images from 10 to $10^5$. On the other hand, for cross-attention approaches SGRAF and NAAF, when a text query arrives, these methods have to pair the text query with every image embedding in the database, then forward each pair through the cross-attention module in order to calculate their similarity. \cref{fig:inference_time} shows that the inference time for SGRAF and NAAF scale up linearly w.r.t. the number of images in the database (\eg, SGRAF takes 46s with $10^4$ images, and 470s with $10^5$ images), which is due to the iterative pairing of the input text with each image. Our model CORA enjoys the benefit of being fast and scalable of the dual-encoder architecture, while still achieving better retrieval results than SOTA cross-attention approaches (\eg, SGRAF and NAAF).

\section{More Ablation Studies}
\label{sec:more_ablation}
\mypara{Initialize from GloVe vs. from scratch} When using Bi-GRU, we follow all recent studies~\cite{long2022gradual,wang2022coder,zhang2022negative,pan2023fine,kim2023improving,fu2023learning} to initialize the word embeddings using GloVe~\cite{pennington2014glove}. To fairly compare against other methods prior to these work, we also report our results when using Bi-GRU with word embeddings initialized from scratch in~\cref{tab:f30_glove_scratch,tab:coco_glove_scratch} for the Flickr30K and MS-COCO dataset respectively. The results show that even when initializing the word embeddings from scratch, our method CORA still outperforms all previous work with and without cross-attention.

\begin{table}
\centering
\caption{\textbf{Our framework achieves the best results on the Flickr30K dataset when initializing the word embeddings fom scratch for the Bi-GRU semantic encoder.} Without the CA - ``cross-attention'', our method still has competitive results to other baselines. $\dagger$ denotes methods that use ensembling of multiple models, and we highlight \textbf{the highest} and \underline{second-highest} RSUM.}
\scalebox{0.6}{
    \begin{tabular}{@{}l|c|c|ccc|ccc|c@{}}
    \toprule
    
    \multirow{2}{*}{Method}
    & \multirow{2}{*}{Venue}
    & \multirow{2}{*}{CA}
    & \multicolumn{3}{c|}{Image $\rightarrow$ Text} 
    & \multicolumn{3}{c|}{Text $\rightarrow$ Image} 
    & \multirow{2}{*}{RSUM} \\
    \cmidrule{4-9}
    
    & & & R@1 & R@5 & R@10 & R@1 & R@5 & R@10 & \\ \midrule
    
    \multicolumn{10}{l}{\textit{\textbf{Faster R-CNN + Bi-GRU}}} \\ \midrule

    SCAN$^\dagger$~\cite{lee2018stacked} & ECCV'18 & \cmark & 67.4& 90.3& 95.8& 48.6& 77.7& 85.2& 465.0 \\

    VSRN~\cite{li2019visual} & ICCV'19 & & 71.3& 90.6& 96.0& 54.7& 81.8& 88.2& 482.6 \\

    SGM~\cite{wang2020cross} & WACV'20 & \cmark & 71.8 & 91.7 & 95.5 & 53.5 & 79.6 & 86.5 & 478.6 \\
    
    GCN+DIST~\cite{li2020visual} & CVPR'20 & \cmark & 70.8 & 92.7 & 96.0 & 60.9 & 86.1 & 91.0 & 497.5 \\

    GSMN$^\dagger$~\cite{liu2020graph} & CVPR'20 & \cmark & 76.4 & 94.3 & 97.3 & 57.4 & 82.3 & 89.0 & 496.8 \\

    CAAN~\cite{zhang2020context} & CVPR'20 & \cmark & 70.1 & 91.6 & 97.2 & 52.8 & 79.0 & 87.9 &  478.6 \\

    VSE${_\infty}$~\cite{chen2021learning} & CVPR'21 & & 76.5& 94.2& 97.7& 56.4& 83.4& 89.9& 498.1 \\

    SGARF$^\dagger$~\cite{diao2021similarity} & AAAI'21 & \cmark & 77.8 & 94.1 & 97.4 & 58.5 & 83.0 & 88.8 & 499.6 \\

    MV-VSE$^\dagger$~\cite{li2022multi} & IJCAI'22 & & 79.0 & 94.9 & 97.7 & 59.1 & 84.6 & 90.6 & 505.8 \\

    \rowcolor{Gray}
    \textbf{Ours} & & & 80.1 & 95.5 & 97.7 & 60.6 & 85.6 & 91.1 &\underline{510.5} \\
    \rowcolor{Gray}
    \textbf{Ours}$^\dagger$ & & & 81.7 & 95.5 & 98.1 & 62.0 & 86.6 & 91.8 & \textbf{515.7} \\

    \bottomrule
    \end{tabular}
}

\vspace{-2mm}
\label{tab:f30_glove_scratch}
\end{table}

\begin{table*}[t]
    \centering
    \caption{\textbf{Our framework achieves the best results on the MS-COCO dataset when initializing the word embeddings fom scratch for the Bi-GRU semantic encoder.} Without the CA - ``cross-attention'', our method still has competitive results to other baselines. $\dagger$ denotes methods that use ensembling of multiple models, and we highlight \textbf{the highest} and \underline{second-highest} RSUM.}
    \scalebox{0.88}{
    \footnotesize
    \setlength{\tabcolsep}{5pt}
    \begin{tabular}{@{}l|c|c|ccc|ccc|c|ccc|ccc|c@{}}
    \toprule
        \multicolumn{1}{l}{\multirow{3}{*}[-1.3em]{Method}}
        & \multicolumn{1}{l}{\multirow{3}{*}[-1.3em]{\ \ \ Venue}}
        & \multicolumn{1}{c|}{\multirow{3}{*}[-0.65em]{Cross-}} 
        & \multicolumn{7}{c|}{MS-COCO 5-fold 1K Test} 
        & \multicolumn{7}{c}{MS-COCO 5K Test} \\ \midrule
        & & 
        & \multicolumn{3}{c|}{Image $\rightarrow$ Text} 
        & \multicolumn{3}{c|}{Text $\rightarrow$ Image}
        & \multicolumn{1}{c|}{\multirow{2}{*}{\textsc{RSUM}}}
        & \multicolumn{3}{c|}{Image $\rightarrow$ Text} 
        & \multicolumn{3}{c|}{Text $\rightarrow$ Image}
        & \multicolumn{1}{c}{\multirow{2}{*}{RSUM}} \\
        \cmidrule{4-9} \cmidrule{11-16}
        \multicolumn{1}{c|}{}
        & & Attention
        & R@1 & R@5 & R@10 & R@1 & R@5 & R@10 &
        & R@1 & R@5 & R@10 & R@1 & R@5 & R@10 & \\
    \midrule %
    \multicolumn{16}{l}{\textit{\textbf{Faster R-CNN + Bi-GRU}}} \\ \midrule
    
    SCAN$^\dagger$~\cite{lee2018stacked} & ECCV'18 & \cmark
        & 72.7& 94.8& 98.4& 58.8& 88.4& 94.8& 507.9
        & 50.4& 82.2& 90.0& 38.6& 69.3& 80.4& 410.9 \\
    
    VSRN~\cite{li2019visual} & ICCV'19 &
        & 76.2 & 94.8 & 98.2 & 62.8 & 89.7 & 95.1 & 516.8
        & 53.0 & 81.1 & 89.4 & 40.5 & 70.6 & 81.1 & 415.7 \\
    
    SGM~\cite{wang2020cross} & WACV'20 & \cmark
        & 73.4 & 93.8 & 97.8 & 57.5 & 87.3 & 94.3 & 504.1 
        & 50.0 & 79.3 & 87.9 & 35.3 & 64.9 & 76.5 & 393.9 \\
    
    CAAN~\cite{zhang2020context} & CVPR'20 & \cmark
        & 75.5 & 95.4 & 98.5 & 61.3 & 89.7 & 95.2 & 515.6
        & 52.5 & 83.3 & 90.9 & 41.2 & 70.3 & 82.9 & 421.1 \\
    
    VSE$_\infty$~\cite{chen2021learning} & CVPR'21 &
        & 78.5 & 96.0 & 98.7 & 61.7 & 90.3 & 95.6 & 520.8 
        & 56.6 & 83.6 & 91.4 & 39.3 & 69.9 & 81.1 & 421.9 \\
    
    SGARF$^\dagger$~\cite{diao2021similarity} & AAAI'21 & \cmark
        & 79.6 & 96.2 & 98.5 & 63.2 & 90.7 & 96.1 & 524.3 
        & 57.8 & - & 91.6 & 41.9 & - & 81.3 & - \\
    
    MV-VSE$^\dagger$~\cite{li2022multi} & IJCAI'22 & 
        & 78.7 & 95.7 & 98.7 & 62.7 & 90.4 & 95.7 & 521.9
        & 56.7 & 84.1 & 91.4 & 40.3 & 70.6 & 81.6 & 424.6 \\
    
    \rowcolor{Gray}
    \textbf{Ours} & & & 80.5 & 96.0 & 98.6 & 62.9 & 90.6 & 96.0 & \underline{524.6} 
        & 60.4 & 85.1 & 91.7 & 40.5 & 70.6 & 81.2 & \underline{429.5} \\
    \rowcolor{Gray}
    \textbf{Ours}$^\dagger$ & & & 81.2 & 96.2 & 98.7 & 63.4 & 90.9 & 96.2 & \textbf{526.6}
        & 60.9 & 85.6 & 92.0 & 40.8 & 71.0 & 81.8 & \textbf{432.1} \\

    \bottomrule %
\end{tabular}
}
\vspace{1mm}

\label{tab:coco_glove_scratch} 
\end{table*}

\begin{table}[t]
\centering
\caption{
\textbf{Ablation studies} to compare between fine-tuning the whole BERT model versus using P-Tuning v2~\cite{liu2021p} to encode the short phrases of semantic concepts. The models are evaluated on the MS-COCO 1K Test set. Gray denotes our best model.
}
\scalebox{0.7}{
    \begin{tabular}{@{}l|ccc|ccc|c@{}}
    \toprule
    
    \multirow{2}{*}{Method}
    & \multicolumn{3}{c|}{Image-to-text} 
    & \multicolumn{3}{c|}{Text-to-image} 
    & \multirow{2}{*}{RSUM} \\
    
    & R@1 & R@5 & R@10 & R@1 & R@5 & R@10 & \\ \midrule
    
    BERT & 82.0 & 96.5 & 98.8 & 64.5 & 91.1 & 96.2 & 529.1 \\

    BERT P-Tuning, N = 8 & 81.7 & 96.5 & 99.0 & 64.5 & 91.1 & 96.1 & 528.9 \\

    BERT P-Tuning, N = 16 & 81.4 & 96.9 & 98.8 & 65.0 & 91.2 & 96.3 & 529.6 \\

    \rowcolor{Gray}
    BERT P-Tuning, N = 24 & 81.9 & 96.6 & 98.9 & 65.0 & 91.2 & 96.4 & 530.0 \\

    BERT P-Tuning, N = 32 & 82.2 & 96.7 & 98.7 & 64.8 & 91.3 & 96.2 & 529.9 \\

    \bottomrule
    \end{tabular}
}
\vspace{0mm}
\label{tab:ablation_ptuning}
\end{table}

\mypara{BERT P-Tuning v2} We compare between direct fine-tuning the whole BERT model against using P-Tuning v2~\cite{liu2021p} to encode short phrases of semantic concepts. The results are displayed in~\cref{tab:ablation_ptuning}. Note that this model is ablated without having multi-head self-attention in the visual encoder.

\section{More Analysis}
\label{sec:more_analysis}

\begin{table}[t]
\centering
\caption{Analysis on using the larger visual backbone ResNeXT-101~\cite{xie2017aggregated}. We plug ResNeXT-101 into our CORA model with BERT as the semantic concept encoder.}
\setlength{\tabcolsep}{5.pt}
\scalebox{0.65}{
    \begin{tabular}{@{}l|ccc|ccc|c@{}}
    \toprule

    \multirow{2}{*}{Method}
    & \multicolumn{3}{c|}{Image-to-text} 
    & \multicolumn{3}{c|}{Text-to-image} 
    & \multirow{2}{*}{RSUM} \\
    
    & R@1 & R@5 & R@10 & R@1 & R@5 & R@10 & \\ \midrule

    CORA-BERT & 84.4 & 96.7 & 98.7 & 62.3 & 87.5 & 92.5 & 522.1 \\

    \rowcolor{Gray}CORA-BERT + ResNeXT-101 & 90.9 & 99.1 & 99.8 & 76.8 & 95.1 & 97.7 & 559.4 \\

    VSE$_\infty$ + ResNeXT-101 [4] & 88.7 & 98.9 & 99.8 & 76.1 & 94.5 & 97.1 & 555.1 \\

    SDE + ResNeXT-101 [16] & 90.6 & 99.0 & 99.6 & 75.9 & 94.7 & 97.3 & 557.1 \\
    
    \bottomrule
    \end{tabular}
}
\vspace{0mm}
\label{tab:resnext} 
\end{table}

\mypara{With larger visual backbone} We select ResNeXT-101~\cite{xie2017aggregated} pretrained on the Instagram dataset~\cite{mahajan2018exploring} as the larger visual extractor than the Faster R-CNN model used in our main experiments. This visual backbone is also reported in VSE$_\infty$~\cite{chen2021learning} and SDE~\cite{kim2023improving}. The results of this experiment on the Flickr30K test set are displayed in~\cref{tab:resnext}, where it shows we obtain a large increase over region features and others.

\mypara{Simulate parsing errors} As discussed in the conclusions, our CORA model is strongly dependent on the scene graph quality from the parser. To study this dependence, we simulate errors by performing the followings onto the parsed graphs: drop word tokens from nodes and edges, move attribute node to wrong object node, and move edge to wrong object pair. We randomly perform these onto 10\%, 20\%, 30\% F30K captions and achieve 513.2, 512.2, 509.8 RSUM (original performance is 515.8). We observe that moving the edge affects performance more than moving the attribute.

\mypara{Why consider CORA} CORA is a promising graph method that can supplement what CLIP (\& other text encoders) may struggle against, \ie, sentences with many semantics that are mixed among objects (discussed in~\cref{sec:intro}). To show example, we select 100 sentences in Flickr30K with the highest number of attributes and relations, then evaluate image retrieval on them. We obtain results respectively for CLIP, HREM, CORA as 239.3, 240.0, 241.5 RSUM. This shows the large model CLIP is even slightly inferior to CORA on sentences that are rich in semantics.

\begin{table}[t]
\centering
\caption{Compare with pre-trained image-text models.}
\setlength{\tabcolsep}{5.pt}
\scalebox{0.85}{
    \begin{tabular}{@{}l|c@{}}
    \toprule

    \multirow{1}{*}{Method}
    & \multirow{1}{*}{RSUM} \\ \midrule

    \multicolumn{1}{l}{ViLBERT - NeurIPS'19 - Data: CC3M} & 502.7 \\

    \multicolumn{1}{l}{UNITER - ECCV'20 - Data: CC3M, SBU} & 510.9 \\

    \multicolumn{1}{l}{UNITER - ECCV'20 - Data: CC3M, SBU, VG, COCO} & 542.8 \\
    
    \multicolumn{1}{l}{Unicoder-VL - AAAI'20 - Data: CC3M, SBU} & 538.8 \\
    
    \rowcolor{Gray}\multicolumn{1}{l}{CORA-BERT - Data: CC1M} & 530.1 \\
    
    \multicolumn{1}{l}{CLIP \underline{zero-shot} - ICML'21 - Data: CLIP 400M} & 540.6 \\
    
    \bottomrule
    \end{tabular}
}
\vspace{-3mm}
\label{tab:image_text_model} 
\end{table}

\mypara{Compare with pretrained image-text models} We scale up CORA on larger data and compare with prior SOTA image-text models in~\cref{tab:image_text_model}. We pretrain CORA on 1M image-text pairs in Conceptual Captions. All of the models in the table are finally fine-tuned on Flickr30K. Compared with CORA-BERT (refer to~\cref{tab:f30k_comparison}), CORA pre-trained gets a +6.8 score. Despite smaller data and not using cross-attention, CORA is better than ViLBERT~\cite{lu2019vilbert}, UNITER~\cite{chen2020uniter}, and can potentially reach Unicoder~\cite{huang2019unicoder} with more data. However, it is inferior to CLIP zero-shot~\cite{radford2021learning}. This shows the promising ability to scale up CORA further, \eg by using ViT instead of regions, CLIP-text instead of BERT, and more data.

\section{Qualitative Results}
\begin{figure*}[t]
\centering
\includegraphics[width=0.88\linewidth]{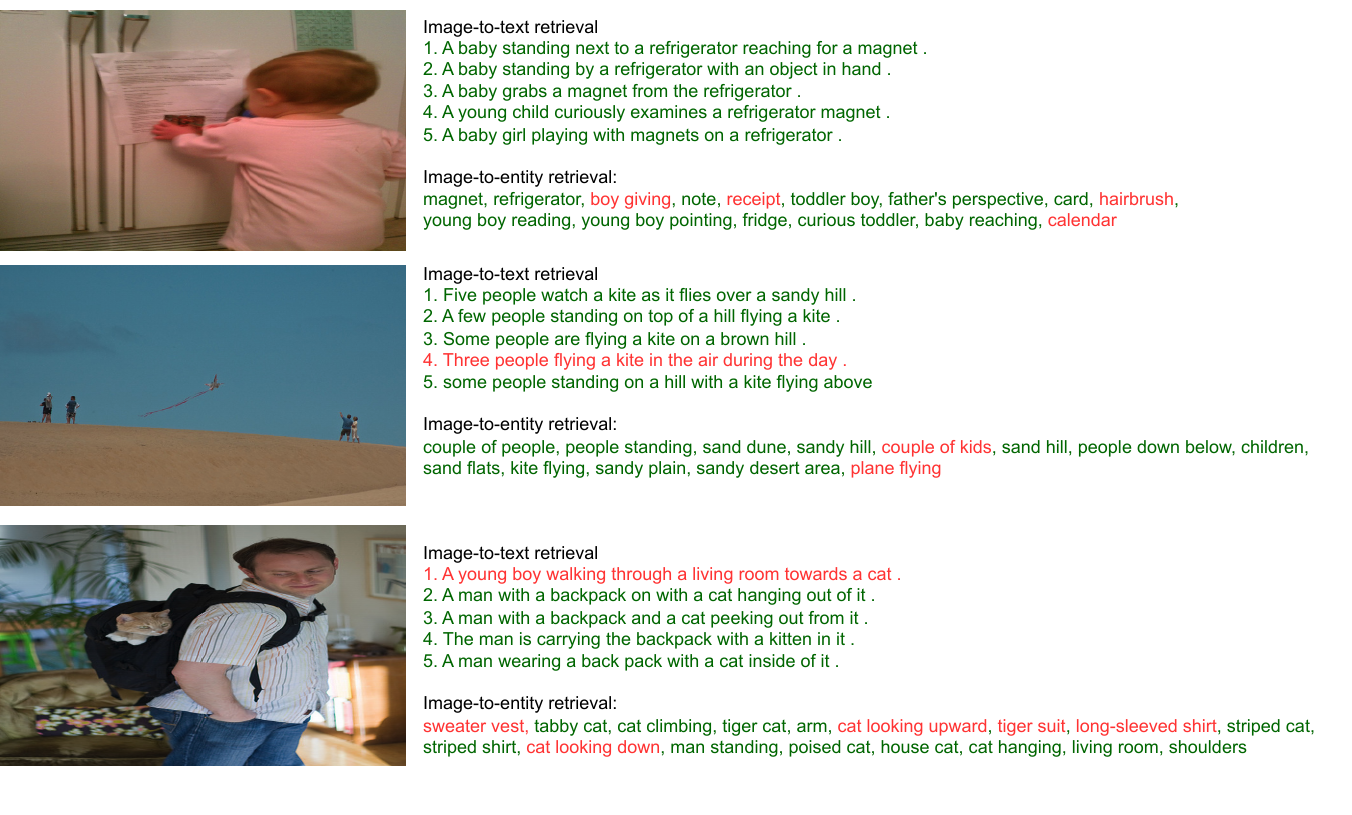}
    \caption{\textbf{Successful image-to-text and image-to-entity retrieval on MS-COCO.} In image-to-text retrieval, green denotes matching text according to the ground truth of MS-COCO, while red denotes incorrect matching. In image-to-entity retrieval, green and red denote correct and incorrect matching, respectively, as judged subjectively by us.}
\label{fig:img2txt_succ}
\end{figure*}

\begin{figure*}[h]
\centering
\includegraphics[width=0.88\linewidth]{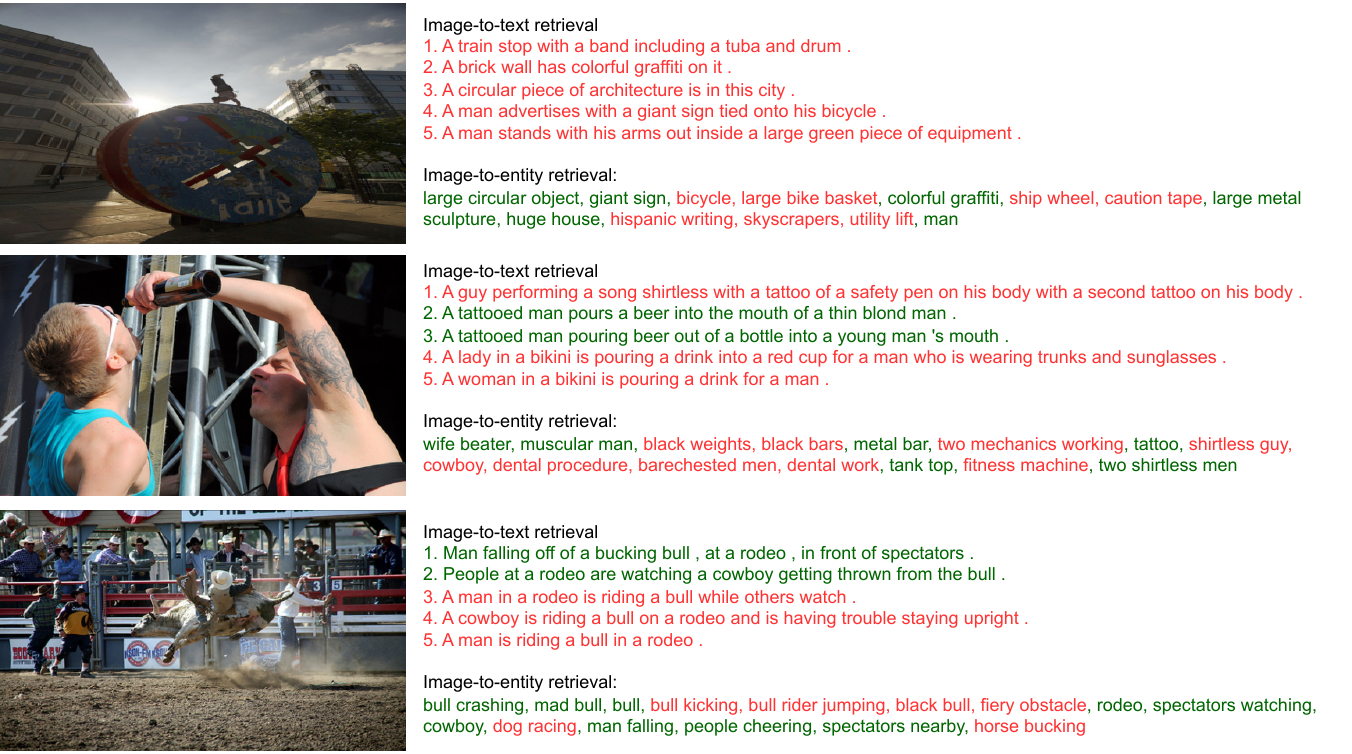}
    \caption{\textbf{Failure cases of image-to-text and image-to-entity retrieval on Flickr30K.} In image-to-text retrieval, green denotes matching text according to the ground truth of Flickr30K, while red denotes incorrect matching. In image-to-entity retrieval, green and red denote correct and incorrect matching, respectively, as judged subjectively by us.}
\label{fig:img2txt_fail}
\end{figure*}

\mypara{Image-to-text and image-to-entity retrieval} We illustrate some examples of successful and failed results when performing image-to-text retrieval using our CORA model with Faster R-CNN + BERT trained on the MS-COCO dataset in~\cref{fig:img2txt_succ} and~\cref{fig:img2txt_fail}. Because CORA also has the ability to retrieve object entities, we also include image-to-entity retrieval results in the figures. The image-to-entity retrieval results also help display some of the biases of the model. One interesting application of image-to-entiy retrieval is for auto image tagging.

Among the examples in~\cref{fig:img2txt_succ}, the wrong matching texts and entities are understandable because they are still very semantically aligned with the input image. We explain each case below:
\begin{enumerate}
    \item In the top image, all retrieved captions are correct. Among the retrieved entities, there are a few incorrect results which show that the model has not learned very accurately the visual appearance of \textit{receipt, hairbrush, calendar}. Images of \textit{toddler holding a hairbrush} is common in the training set, which must have made the model steered towards aligning \textit{hairbrush} with something that \textit{a toddler is holding}.
    \item In the middle image, most matching captions are correctly retrieved except one that is incorrect due to object counting. Counting the correct number of objects is indeed a challenge for image-text matching model. The model also mistakenly recognizes the kite as a plane.
    \item In the bottom image, the 1st caption is incorrect, but the model still ranks it at the top due to multiple semantic information in the text are still correct w.r.t. the image (\eg, \textit{young boy, living room, cat}). All other captions are correctly retrieved. The image-to-entity retrievals show the concepts that the model does not grasp well.
\end{enumerate}

We continue to explain the failure cases in~\cref{fig:img2txt_fail} as following:
\begin{enumerate}
    \item In the top image, all of the retrieved captions are incorrect matchings as determined by the ground truth data. However, we notice that the 2nd, 3rd captions still correctly describe the image to a certain extent. This is a weakness of the benchmark.
    \item In the middle image, the model must have wrongly associated \textit{skin} with \textit{bikini}, hence why it retrieves captions with \textit{bikini} at rank 4 and 5. In the entity retrieval results, interestingly, we notice the model returns \textit{dental procedure} and \textit{dental work}. We figure that the model must have aligned the action of \textit{mouth opening} with \textit{dental}, hence why these two entities are retrieved in this case.
    \item In the bottom image, this is again an example of where the retrieved captions correctly describe the image, but because the ground truth data specify otherwise, they are considered incorrect by the benchmark.
\end{enumerate}

\begin{figure*}
\centering
\includegraphics[width=0.93\linewidth]{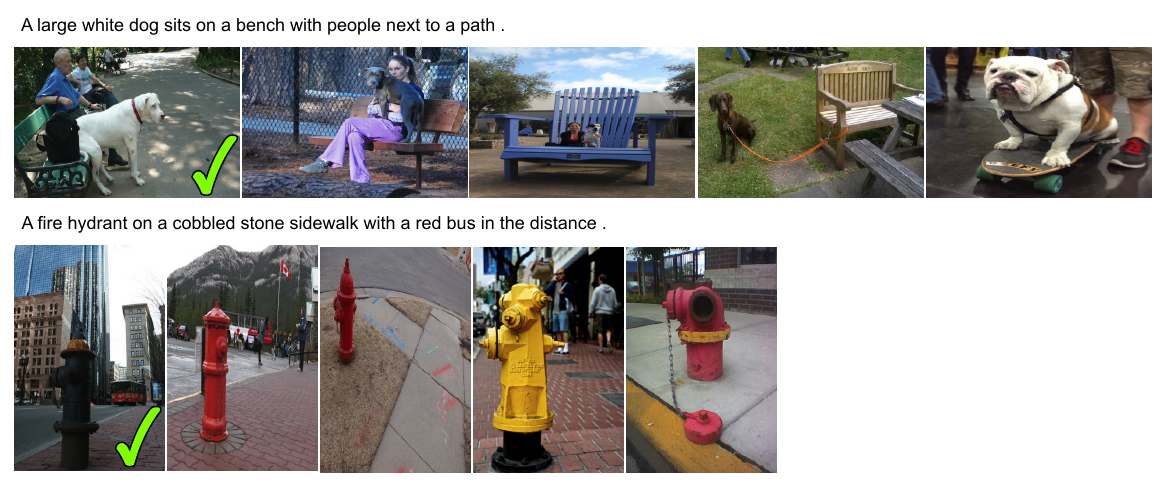}
    \caption{\textbf{Text-to-image retrieval on MS-COCO.} For every text, we show the top-5 retrieved images on MS-COCO. The image with the green tick mark is the correct matching according to ground truth in the dataset.}
\label{fig:txt2img}
\end{figure*}

\mypara{Text-to-image retrieval results} We illustrate some text-to-image retrieval results in~\cref{fig:txt2img}. In both examples, our model is able to retrieve the correct image at rank 1. The images from rank 2 to rank 5 all exhibit visual traits that match partially with the input text.

\end{document}